\documentclass[12pt, a4paper]{article}
\usepackage[authoryear]{natbib}

\usepackage{setspace}
\usepackage{style}

\usepackage{lastpage}
\jmlrheading{23}{2022}{1-\pageref{LastPage}}{1/21; Revised 5/22}{9/22}{21-0000}{Author One and Author Two}

\ShortHeadings{AIA Forecaster: Technical Report}{AIA Labs}
\firstpageno{1}

\usepackage[dvipsnames]{xcolor}  % Enables more color names like NavyBlue, Maroon, etc.

\usepackage{hyperref}
\hypersetup{
    colorlinks=true,
    linkcolor=darkblue,
    citecolor=darkblue,
    filecolor=darkblue,
    urlcolor=darkblue,
    bookmarksopen=true,
    bookmarksnumbered=true,
    pdfborder={0 0 0}
}

\usepackage[utf8]{inputenc}
\usepackage[T1]{fontenc}
\usepackage{amsmath, amssymb, amsfonts}
\usepackage{graphicx}
\usepackage[margin=1in]{geometry}
\usepackage{booktabs}

\usepackage{pdflscape}  % For landscape environment
\usepackage{afterpage}  % For better page break handling (optional but recommended)

\usepackage{lmodern}
\usepackage{microtype}

\usepackage[table]{xcolor}
\usepackage{colortbl}
\definecolor{darkblue}{RGB}{20,56,126}
\definecolor{lightgray}{RGB}{240,240,240}
\definecolor{mediumgray}{RGB}{128,128,128}

\usepackage[table]{xcolor}   % for \columncolor
\usepackage{tcolorbox}       % for colored/framed boxes
\tcbuselibrary{skins}        % optional skins enhancements
\usepackage[section]{placeins} % auto-places a barrier at each \section

\usepackage{array}
\usepackage{multirow}
\usepackage{siunitx}
\usepackage{listings}
\usepackage{comment}
\usepackage{xspace}
\usepackage{multicol}
\usepackage{enumitem}

\usepackage{color-edits}
\addauthor[RA]{ra}{blue}
\addauthor[MM]{mm}{green!60!black}
\addauthor[RC]{rc}{red!60!black}

\usepackage{tocloft}

\usepackage[table]{xcolor}   % for \columncolor
\usepackage{tcolorbox}       % for the framed box
\tcbuselibrary{skins}        % optional tcolorbox tweaks
\usepackage{xstring}         % for \IfStrEq
\usepackage{xparse}   
\usepackage[justification=centering,singlelinecheck=false]{caption}

\newcommand{\systemname}{\textsc{AIA Forecaster}\xspace}
\newcommand{\forecastbench}{\textsc{ForecastBench}\xspace}
\newcommand{\MarketLiquid}{\textsc{MarketLiquid}\xspace}
\newcommand{\MarketNightly}{\textsc{MarketNightly}\xspace}
\newcommand{\fbOld}{\textsc{FB-7-21}\xspace}
\newcommand{\fbOldMarket}{\textsc{FB-Market}\xspace}
\newcommand{\fbNew}{\textsc{FB-8-14}\xspace}

\newcommand{\ProviderA}{\textsc{Search-A}\xspace}
\newcommand{\ProviderB}{\textsc{Search-B}\xspace}

\begin{document}

\title{AIA Forecaster: Technical Report}

\makeatletter
% Save original footnote definitions
\let\orig@makefnmark\@makefnmark
\let\orig@makefntext\@makefntext

% Redefine footnote formatting to remove the period
\renewcommand\@makefnmark{\hbox{\@textsuperscript{\normalfont*}}}
\renewcommand\@makefntext[1]{\noindent\@makefnmark\ #1}
\makeatother

\author{Rohan Alur$^{*}$, 
        Bradly C.\ Stadie$^*$, 
        Daniel Kang, Ryan Chen, Matt McManus, Michael Rickert, Tyler Lee, Michael Federici, Richard Zhu, Dennis Fogerty, Hayley Williamson, Nina Lozinski, Aaron Linsky, Jasjeet S.\ Sekhon \\ \\
        \addr{Bridgewater AIA Labs \\
        New York, NY}}

% \addtocounter{footnote}{-1}
% \renewcommand{\thefootnote}{\fnsymbol{footnote}}
% \stepcounter{footnote}
% \footnotetext{*These authors contributed equally to this work}
% \renewcommand{\thefootnote}{\arabic{footnote}}

\date{}

\maketitle

% Custom footnote without the period
\footnotetext{These authors contributed equally to this work. Correspondence to aialabs@bwater.com.}

% Restore original footnote definitions
\makeatletter
\let\@makefnmark\orig@makefnmark
\let\@makefntext\orig@makefntext
\makeatother

% \def\thefootnote{*}\footnotetext{These authors contributed equally to this work.}\def\thefootnote{\arabic{footnote}}

% Enhanced title page
\maketitle
\thispagestyle{empty}

\vspace{2cm}

\begin{abstract}%   
This technical report describes the \systemname, a Large Language Model (LLM)-based system for judgmental forecasting using unstructured data. The \systemname approach combines three core elements: agentic search over high-quality news sources, a supervisor agent that reconciles disparate forecasts for the same event, and a set of statistical calibration techniques to counter behavioral biases in large language models. On the \forecastbench benchmark \citep{karger2024forecastbench}, the \systemname achieves performance equal to human superforecasters, surpassing prior LLM baselines. In addition to reporting on \forecastbench, we also introduce a more challenging forecasting benchmark sourced from liquid prediction markets. While the \systemname underperforms market consensus on this benchmark, an ensemble combining \systemname with market consensus outperforms consensus alone, demonstrating that our forecaster provides additive information. Our work establishes a new state of the art in AI forecasting and provides practical, transferable recommendations for future research. To the best of our knowledge, this is the first work that verifiably achieves expert-level forecasting at scale. 

\end{abstract}

\newpage
\tableofcontents
\newpage

\section{Introduction}

Forecasting is a universal problem. Any system that makes informed judgments about what may happen in the future relies on at least some degree of forecasting. In agriculture, this takes the form of predicting future crop yields and food system resilience \citep{bueechi2023crop, tanaka2023satellite, paudel2021machine}. In biotechnology and medicine, clinical trials are carefully designed to help researchers predict the impact of newly developed treatments \citep{qian2025enhancing, curth2024using}. Elsewhere in biology, AlphaFold can be viewed as a form of forecasting protein folding structure under uncertainty \citep{jumper2021highly}. In academia, grant funding is a form of forecasting which scientific discoveries will be made conditional on funding \citep{tohalino2022predicting}. Climate and environmental science models local and long-term weather trends, which are of universal importance \citep{soliman2024deepmind, price2025probabilistic, deepmind2024weathernext}. In political science, forecasting election outcomes and the impact of policy decisions are an essential component of political calculus \citep{jennings2020election, wang2015forecasting, gelman2020information}. Militaries make use of forecasting for threat assessment and conflict predictions \citep{zequeira2024artificial, probasco2025ai}. Forecasting is the very heartbeat of financial firms, which use it to estimate future economic conditions in an uncertain world \citep{tsay2005analysis}. Wherever there is uncertainty about the future, there is forecasting. 

The literature distinguishes between two high-level approaches to forecasting. The first, statistical forecasting, utilizes some combination of tabular data and simulation to develop a mathematical model of what might occur in the future. In contrast, \textit{judgmental forecasting} is the process of aggregating unstructured data (e.g., news articles, scientific reports, etc.) and using this data in conjunction with past experience to logically predict some future outcome \citep{LAWRENCE2006493}.

To better understand the two types of forecasting and their relationship, let us consider an example. Suppose a business wants to predict whether a certain piece of real estate will appreciate in value next year. By comparing the property to other similar properties, and looking at recent trends in the market, the company can develop a statistical model with some degree of sophistication. However, this modeling has its limits. Suppose that a news article is released, claiming that the foundation of a nearby building is shifting, causing it to slowly tilt. Even without performing additional research, a human can predict that this fact will likely put downward pressure on the property value. However, coercing this information into a form amenable to mathematical analysis is difficult. Thus, an ideal forecast would employ some abstract reasoning process which aggregates unstructured data, past experience, and logical thinking (i.e., judgment) to arrive at a sound conclusion.

This paper will focus exclusively on the judgmental forecasting problem (hereafter referred to simply as ``forecasting''). Recently, there has been an uptick in research which studies the judgmental forecasting problem, due in large part to the ubiquity of Large Language Models (LLMs). By giving LLMs access to news databases and asking them to forecast future events, it is possible to systemically evaluate the forecasting abilities of machine intelligence at scale. Several recent papers have endeavored to create and evaluate LLM-based forecasters \citep{halawi2024approaching, karger2024forecastbench, schoenegger2024wisdom, turtel2025llms, schoenegger2025prompt}. The rapid advances of LLMs on reasoning tasks (e.g., achieving gold medal performance at the International Math Olympiad \citep{deepmind2025gemini}) offer reasons for optimism, and prompt the obvious question:

\begin{center}
    \emph{Can LLMs match or even exceed the performance of expert human forecasters?}
\end{center}

\textbf{Contributions.} We answer this question in the affirmative by introducing the \systemname. On \forecastbench \citep{karger2024forecastbench}, the most widely used academic benchmark for forecasting, we achieve results that are statistically indistinguishable from human superforecasters. We also introduce a new benchmark which consists of 1610 questions sourced from a popular public prediction market platform. These questions cover decision-relevant events sourced from liquid markets, and are significantly harder than those on \forecastbench. On this benchmark, the \systemname lags slightly behind market consensus. However, an ensemble of the system's forecasts with the market consensus outperforms consensus alone, indicating that the \systemname is diversifying with respect to market prices. We corroborate these results by establishing a track record of real-time, forward-looking forecasts. In this modest sample, the \systemname is highly competitive with market consensus in liquid prediction markets. 

In addition to our core experiments, we also conduct extensive analysis to understand what makes a good forecaster. Our methodological contributions are as follows:

\begin{enumerate}
\item The \systemname is equipped with the ability to perform a robust search over a large corpus of high-quality documents. Critically, this search is both \emph{agentic} and \emph{adaptive} in nature: the model is granted full discretion over what to search, and can adaptively seek out new information by conditioning on the results of earlier queries. We show there is a deep relationship between search quality and forecast quality, resolving a major open question in the literature.

\item We conduct an exhaustive analysis of \emph{foreknowledge bias}, or unintended violations of the intended information cutoff. In particular, ex-post evaluation of such systems requires the ability to ``travel back in time'' and feed the forecaster news ``as of'' the intended forecasting date. Reliably enforcing this information cutoff is a surprisingly challenging problem \citep{paleka2025pitfalls}. We build a custom pipeline for detecting foreknowledge bias and demonstrate its efficacy, confirming the integrity of our results. We also introduce a live benchmark, \MarketNightly, which includes only active, liquid prediction markets, and thus precludes the use of foreknowledge. Results on this benchmark corroborate our other findings.

\item We show that individual forecast runs are unstable, and that ensembling over many forecasts is absolutely essential to obtain frontier-level forecasting performance. We also show that the exact method of ensembling is crucial. It is not enough to take the simple mean of forecasts. Instead, the \systemname passes the results of individual forecasting agents to a supervisor agent, which performs additional search to correct mistakes and reconcile differing perspectives.

\item LLMs are fundamentally miscalibrated for probabilistic prediction under uncertainty. They hedge too much and are too cautious. We examine a variety of different statistical correction techniques that can be applied to correct these biases. We derive a mathematical connection, which has not yet been documented in the literature, between Platt scaling and the extremization of forecasts. We also show that simple extremization techniques are often sufficient to obtain large gains in forecasting performance.
\end{enumerate}

\textbf{Organization.} In Section \Ref{sec:related}, we provide an overview of related work. In Section \Ref{sec:methods}, we describe the architecture of the \systemname, and in Section \Ref{sec:main} we present our main results on multiple variants of \forecastbench and our novel \MarketLiquid benchmark. In Section \Ref{sec:search} we investigate the role of search, and address methodological challenges related to \emph{foreknowledge bias}. We complement these results in Section \Ref{sec:live_markets}, where we study real-time forecasting performance in live prediction markets. In Section \Ref{sec:synthesis}, we study a range of ensembling and statistical calibration techniques, which we show can substantially improve on the performance of a single forecasting agent. Finally, in Section \Ref{sec:discussion}, we conclude with a discussion of limitations, open questions, and future work.

\section{Related Work}
\label{sec:related}

We focus on \emph{judgmental forecasting}, in which a forecaster is tasked with producing the probability that a binary event will occur \citep{lahiri2013forecasting, tetlock2016superforecasting, tetlock2017expert, hanson1999decision}. In the past 30 years, judgmental forecasting has become an increasingly important area of study \citep{LAWRENCE2006493}. Two major findings that have increased its salience are (1) that the \emph{wisdom of crowds} (aggregated predictions across many forecasters) often beats experts in judgmental forecasting \citep{Surowiecki2004, bassambooWisdom2015, zhi2019wisdom, lichtendahl2013wisdom}, and (2) the existence of \emph{superforecasters}, who consistently outperform even the wisdom of crowds \citep{tetlock2017expert, tetlock2016superforecasting}. The literature demonstrates that superforecasters are rare, and even subject-matter experts are often poor forecasters in their area of expertise. What's more, superforecasting ability in one domain does not necessarily transfer to other domains. 

With the advent of strong AI systems, there has been increased interest in AI forecasting \citep{halawi2024approaching}. AI forecasting is highly scalable, so if the performance of AI forecasters matches or exceeds superforecasters, these systems can be widely deployed. Of course, in adversarial settings (e.g., financial markets), widely available AI forecasters may in turn increase the difficulty of subsequent prediction tasks, requiring systems which continuously evolve.\footnote{More generally, tasks in which predictions influence the distribution of future outcomes are known as \emph{performative} prediction tasks \citep{perdomo2020performative}.} 

\textbf{Techniques for AI forecasters.}
Modern AI forecasters provide context to an LLM and prompt it to produce the probability that a future event will occur \citep{wang2024news, halawi2024approaching, schoenegger2024wisdom}. However, as with any modern AI system, there are many research choices that dramatically affect performance. For example, the LLM's prompt \citep{schoenegger2025prompt} and the news provided to the LLM \citep{wang2024news} have both been shown to carry an outsize impact on AI's forecasting ability \citep{turtel2025llms}.

\textbf{Benchmarks for AI forecasters.}
Recently, several benchmarks have been developed to evaluate the prediction quality of AI forecasters. Broadly, all benchmarks can be partitioned into two categories: \textit{static} and \textit{dynamic}. Static benchmarks are gathered at a single point in time and held fixed indefinitely thereafter. Dynamic benchmarks are continually updated to include fresh questions on a regular basis. 

Examples of static benchmarks include \cite{halawi2024approaching}, along with many others \citep{jin2020forecastqa, mutschlechner2025analyzing, yuan2025future, wang2025openforecast, ye2024mirai, zou2022forecasting}. Static benchmarks will always atrophy as time marches on and new LLMs are trained. This is because newer models have knowledge of the world which is acquired after the question was initially posed. For example, if a benchmark published in 2023 asks the LLM to forecast who will win the 2024 US presidential election, models trained at later dates may acquire knowledge of the outcome (and other closely related events) through their pre-training corpus \citep{sarkar2024lookahead}. Models may also be given access to tools which inadvertently leak knowledge beyond the intended information cutoff \citep{paleka2025pitfalls}. The foreknowledge bias problem, together with several other issues related to the quality of baseline forecasts\footnote{In particular, if one does not collect high-quality superforecaster or crowd predictions at the time the benchmark is created, it is impossible to travel back in time and obtain these baseline forecasts. Many extant benchmarks lack such baselines, which are crucial for evaluation.}, is often the limiting factor in using static benchmarks to evaluate forecasting systems.

Regarding dynamic benchmarks, there are three major choices at present: \forecastbench \citep{karger2024forecastbench}, Prophet Arena \citep{tao2025prophet}, and the Metaculus AI Forecasting Benchmarking Series \citep{metaculus}. \forecastbench sources questions from public prediction markets, including Manifold, Polymarket, and Metaculus. It filters out ``low-quality'' questions using an LLM. Given a list of questions, it then produces ``combination questions'', which ask the model to forecast boolean conjunctions of the original events. Prophet Arena also sources questions from prediction markets, but fixes the news sources the LLMs can use. This simplifies the forecasting task significantly because, as we argue in Section~\ref{sec:search}, effective search is arguably the most critical component of effective forecasting. The Metaculus AI Forecasting Benchmarking Series releases questions on a weekly basis; this benchmark only offers a very small subset (48 questions) on which to evaluate LLMs against human performance.

A major issue with all three dynamic benchmarks is domain-specific relevancy. For example, \forecastbench includes questions on the ELO scores of professional chess players, and the Metaculus AI Forecasting Benchmarking Series includes questions on podcast rankings. To the best of our knowledge, these questions are irrelevant from a policy or economic perspective.

\section{Methodology}
\label{sec:methods}

\subsection{Notation and Preliminaries}

In this work, we consider judgmental binary forecasting, wherein an LLM is presented with some binary question $q$. Toward answering the question $q$ in an informed manner, we allow the LLM to gather some evidence, $\mathcal{E}$, before attempting its answer. For example, we might query an online news database of articles for articles related to $q$. 

Using the gathered evidence, the LLM makes a prediction $p$, which denotes the probability of the outcome under scrutiny occurring. If the LLM is absolutely sure the event in question will come to pass, it should set $p=1.0$. Likewise, if the LLM is sure the event will not happen, it should set $p= 0.0 $. Intermediate values such as $p=0.3$ should be used if the LLM is uncertain about the final outcome, but nonetheless has some evidence which shades its beliefs in a particular direction.

This process of mulling over evidence and assigning a probability can be denoted as 
\begin{align*}
\pi: \left( q, \mathcal{E} \right) \to p 
\end{align*}

Of course, we have hidden a great many difficulties with the explanation above. Principal among them, we supposed the existence of an evidence gathering pipeline $\mathcal{E}$, tasked with collecting information that is useful for informing the LLM’s predictions. As good human researchers know, evidence gathering is often a highly iterative process: some article contains a key piece of information, which sparks further research, cascading in dozens of queries to find the information required to fully understand the underlying situation. Human forecasters may also work collaboratively, exchanging information and engaging in debate to arrive at a final forecast.

More generally, in designing an LLM forecaster, we face a vast array of design decisions, many of which mirror those faced by expert human forecasters. These choices include the choice of retrieval pipeline, the method of reasoning over evidence, and the manner in which to evaluate performance, among others. Below we describe the design of the \systemname, which takes inspiration from the human forecasting process.

\subsection{The \systemname}

\textbf{Architecture.}
The \systemname is a multi-agent forecasting system, in which several agents perform an adaptive search procedure to retrieve relevant information, reason over the retrieved content, and finally synthesize this reasoning to produce a single probability. The system then reconciles these independent forecasts and applies a statistical correction to produce a final answer.

\begin{figure}[htbp]
    \centering
    \includegraphics[width=1\linewidth]{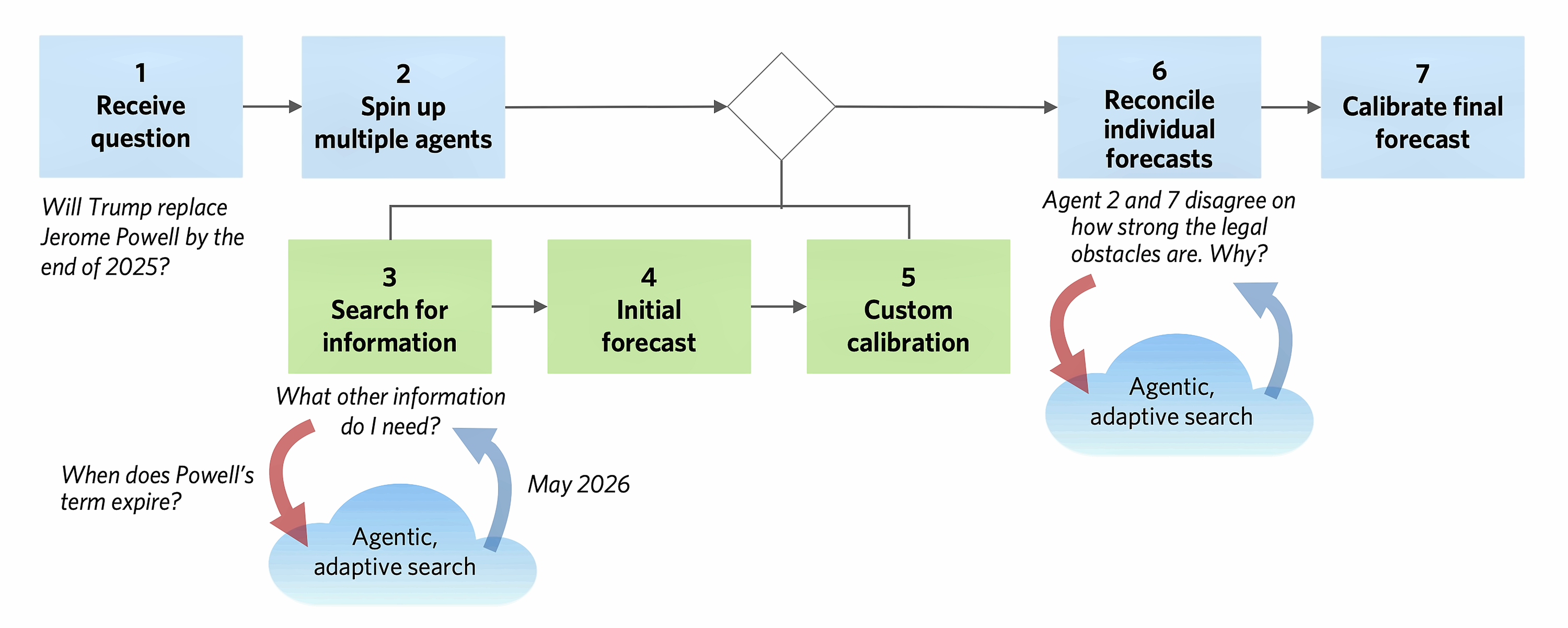}
    \caption{The architecture of the \systemname }
    \label{fig:arch}
\end{figure}

A full visualization of our forecasting pipeline can be found in Figure~\ref{fig:arch}. In the pipeline, we see that $M$ individual agents are spawned. Each agent carries out independent research and produces an initial forecast probability. Those individual forecasts are passed to a supervisor agent, which identifies disagreements among the $M$ agents and is given a final chance to query our news database with any clarifying questions. After clarifying, the reconciliation pipeline outputs an aggregated forecast, which is then passed through a Platt scaling step to produce the final probability.

We now describe the individual components of the \systemname.

\textbf{Search.} Each forecasting agent is granted access to an external search provider, which it can query as needed. As described above, a key feature of this search procedure is that it is \emph{agentic} and \emph{adaptive}: each agent is granted full discretion to determine whether and how to query the search provider, and can iteratively improve the quality of its evidence by conditioning each query on prior results. Each forecasting agent operates independently, and their paths can diverge significantly over the course of a single run. As we discuss below, this can be regarded as both a bug and a feature. We can denote this multi-stage evidence gathering pipeline in mathematical notation as

\begin{align*}
\pi: q \to \mathcal{E}_1 \to \mathcal{E}_2 \to \dots \to  \mathcal{E}_n \to p 
\end{align*}

While agentic search produces far better results than the non-agentic search considered in prior work \citep{halawi2024approaching, karger2024forecastbench}, this change is insufficient to achieve superforecasting levels. There are multiple reasons for this, all of which can be distilled into the fact that individual forecast runs are unstable, and this instability effectively destroys the fidelity of aggregated benchmark results. The instability itself comes from many sources: missing a key piece of information, over-anchoring to an incorrect base rate, and the inherent randomness of LLMs leading to the occasional outlier opinion.

\textbf{Reconciliation.}
One naive solution for addressing run-over-run instability is to simply execute forecasts many times from scratch and to ensemble the results. We show that taking a naive mean over $N$ forecasts is a very strong baseline, and that many advanced ensembling methods fail to outperform this comparatively simple strategy. 

In particular, suppose we try to beat the mean by asking an LLM to carefully read all the individual forecasts, evaluate their relative merits, and then produce a final estimate (as proposed in \cite{halawi2024approaching}). When we do this, we see somewhat surprisingly that the resulting aggregated prediction is substantially less accurate than a simple average of predictions. By inspecting the reasoning traces, we find that LLMs are prone to overemphasizing outlier opinions. This failure exacerbates the instability inherent to LLM forecasting, putting us in a difficult position. 

However, it turns out that a simple reframing of this problem is enough to turn our fortunes around. Rather than asking a supervisor LLM to aggregate all forecasts directly, we instead instruct a supervisor agent to examine all forecasts' reasoning traces $\mathcal{R}_i$ and identify sources of ambiguity or disagreement. This supervisor is allowed to output $N$ search queries, with the hopes that these will help reconcile the differences between competing forecasts. Useful queries at this stage often involve looking up a base rate, or fact-checking assertions made by individual forecasting agents. 

Thus, if we denote information gathering agent $i$ as $\pi_i$, our fully realized evidence gathering pipeline can be written as 

\begin{align*}
\pi_i : q \to \mathcal{E}_1 \to \mathcal{E}_2 \to \dots \to  \mathcal{E}_n \to \left( \mathcal{R}_i, p_i \right) \\
\textrm{Supervisor}:   \left( \mathcal{R}_1, \mathcal{R}_2, \dots, \mathcal{R}_M \right) \to \mathcal{E}_{\textrm{supervisor}} \to p_{\text{final}} 
\end{align*}

\textbf{Hedging.}
Finally, we have one more piece of drama to address. It turns out that LLMs have an annoying bias, likely stemming from post-training via reinforcement learning from human feedback (RLHF) \citep{kaufmann2023rlhf}, which makes them hedge toward 0.5 when making predictions. Forecasts for events where the outcome is somewhat certain (e.g., a true probability of 0.85) are often attenuated to a more entropic forecast (e.g., 0.6). Additionally, even for questions where the outcome is certain, LLMs will often predict 0.95 rather than 1.0, justifying their response with reasoning like ``unexpected things might happen, and we need to be safe.'' We document some examples in Appendix \ref{app:hedging}. These pitfalls can be corrected rather easily with statistical corrections such as Platt scaling. The core idea is to extremize the forecasts \citep{tetlock-extreme-paper}, correcting the output probabilities along a sigmoid curve. In practice, this process pushes low probabilities further down as they get closer to 0.0 and high probabilities higher up as they get closer to 1.0. We explore popular statistical calibration methods, and demonstrate that Platt scaling and a certain class of extremization techniques are mathematically equivalent. Overall, we see that Platt scaling is an essential statistical correction technique for achieving the best possible performance.\footnote{Interestingly, we consistently found this class of mathematical techniques to be more robust than simple prompting changes, which we found largely ineffective at mitigating this problem.}

We now turn to our primary results.

\section{The \systemname Is Indistinguishable from Superforecasters}
\label{sec:main}

\subsection{Benchmarks and Evaluation Metrics}

We report the performance of the \systemname on three variants of the \forecastbench benchmark \citep{karger2024forecastbench}, as well as an additional benchmark (the \MarketLiquid benchmark), which we describe below. Each of these benchmarks covers a diverse set of binary questions spanning macroeconomics, politics, technological developments, sports, and other miscellaneous events. For example, a representative question in \forecastbench asks: 

\begin{center}
    \emph{Will an infrastructure disaster costing $>$\$1B in a G20 country be widely attributed to an AI cyberattack before 2025?}
\end{center}

 Each event in these benchmarks resolves between July 2024 and June 2025, allowing us to observe the ground truth outcome (whether or not the event occurred).

\begin{table}[htbp]
\small
\centering
\begin{tabular}{lrrp{7cm}}
\toprule
\textbf{Name} & \textbf{Questions} & \textbf{Resolution Range} & \textbf{Description} \\
\midrule 
\fbOld & 498 & 7/28/2024 -- 12/31/2050 & Original \citet{karger2024forecastbench} benchmark. \\
\fbNew & 602 & 7/28/2024 -- 12/31/2050  & Updated \citet{karger2024forecastbench} benchmark.\\
\fbOldMarket & 76 & 7/25/2025 -- 12/31/2050 & Subset of \citet{karger2024forecastbench} sourced from prediction markets.\\
\MarketLiquid & 1610 & 4/2/2025 -- 5/23/2025 & More challenging benchmark sourced from liquid prediction markets. \\
\bottomrule
\end{tabular}
\caption{Overview of benchmarks used to evaluate the performance of the \systemname.}
\label{tab:dataset-des} 
\end{table}

\textbf{\forecastbench.} The first variant of \forecastbench, which we call \fbOld, is the 498-question benchmark previously studied in \cite{karger2024forecastbench}. The authors report performance for a wide variety of LLM-based forecasters on this benchmark, as well as the aggregate performance of human-generated forecasts in both general public surveys and surveys of expert superforecasters.\footnote{\cite{karger2024forecastbench} also publish a larger ``LLM-only'' benchmark, which includes synthetically generated conjunctions of the original questions. We focus here on the benchmark which includes human performance.} These questions are collected from a variety of public prediction markets and other established datasets which track world events. For additional detail on the data collection process we refer to \cite{karger2024forecastbench}. 

We also separately report performance on two variants of \forecastbench. The first, \fbOldMarket, is the subset of \fbOld which is sourced from public prediction markets.\footnote{As shown in Table \ref{tab:dataset-des}, the \fbOldMarket dataset includes questions with resolution dates as late as December 31, 2050. Additionally, a few questions are event-contingent and resolve immediately when the specified event occurs, provided it happens before a designated cutoff date (e.g., year 2100). For these questions, the resolution dates are indeed set at a far future date. Thus, the data set accommodates both fixed-date predictions and open-ended event forecasts.} The second, which we call \fbNew, is an updated snapshot of \forecastbench as of 08/14/2025 containing 602 questions.

\textbf{\MarketLiquid.}
Next, we curate a timely dataset of binary forecasting events sourced from a popular public prediction market platform. Our goal was to create a dataset such that (1) all of the markets resolve beyond the knowledge cutoff for the models considered in this manuscript and (2) includes events which are more relevant for policymakers and executives. For example, we restrict our attention to markets covering politics and policy, economics and financial markets, and developments in AI and technology; we exclude markets related to sports, weather prediction, and other miscellaneous categories which appear frequently in \forecastbench.

This dataset includes all markets which resolved between 4/2/2025 and 5/23/2025, but excludes markets which (1) were open for less than one week and (2) had fewer than 5,000 total contracts traded. 
This filtering yields 322 forecasting questions. For each question, we generate binary event forecasting questions at five dates spaced between the posted market open and market resolution date for a total of 1610 questions.\footnote{Note that the market close date (the posted date by which the market will resolve) and the realized resolution date may be distinct, e.g., for events of the form ``Will X happen by Y closing date'' that occur strictly before the closing date. While we generate questions uniformly between the market open and \emph{realized} resolution date, we only disclose the originally posted close date to the LLM forecaster.} We provide additional detail on the curation of this benchmark, including the choice of dates and the prompts used to convert the raw data into a structured benchmark, in Appendix \Ref{app:market}.

\textbf{Evaluation criteria.} As is standard in the LLM forecasting literature (e.g., \cite{halawi2024approaching,karger2024forecastbench,schoenegger2024wisdom}), we use the Brier score to evaluate forecasting performance \citep{brier1950verification} on each of the benchmarks described above. The Brier score for an individual forecast $p \in [0, 1]$ with ground truth outcome $o \in \{0, 1\}$ is simply the squared difference of the two, $(p - o)^2$. The Brier score for a set of events indexed by $i = 1, \dots, n$ is the mean of this quantity, $\frac{1}{n} \sum_{i=1}^n (p_i - o_i)^2$. The Brier score ranges from 0 (perfect accuracy) to 1; a forecaster who uniformly predicts $p = 0.5$ for every event can achieve a baseline score of $0.25$. The Brier score is known to be a strictly proper scoring rule, which incentivizes truthful forecasting (see e.g., \cite{gneiting2007proper} for an overview).

\subsection{Forecaster Performance}

In this section we present our headline results on \fbOldMarket, \fbOld, \fbNew and the \MarketLiquid benchmarks. For \fbOld, we include the performance of the following three baselines published in \cite{karger2024forecastbench}: the state-of-the-art LLM forecaster on the \cite{karger2024forecastbench} leaderboard, the median forecast for each event in a general public survey, and the median forecast for each event in a survey of expert superforecasters.\footnote{We follow \cite{karger2024forecastbench} in using the median forecast for each event as the corresponding baseline for both superforecasters and the general public survey. This choice can be motivated by arguments which demonstrate that, under the assumption that individual forecasters make symmetric errors around the ``true'' forecast, the median is an unbiased estimate of this unknown value but the mean is not \citep{tetlock-extreme-paper}. Our empirical evidence corroborates this intuition: the per-event median of superforecasters (0.1110) modestly outperforms the per-event mean (0.1123), and the per-event median of the general public survey (0.1450) substantially outperforms the per-event mean (0.1586). The general public survey averages 49 responses per question (with a minimum of 40). The survey of superforecasters includes an average of eight responses per question (with a minimum of three). } We also include the performance of OpenAI's o3 reasoning model, which we find is the best off-the-shelf forecaster for which these questions do not appear in its pre-training corpus (see Appendix~\ref{app:base_models} for details). On \fbOldMarket, we further evaluate performance against the market consensus (see Section~\ref{sec:search} for details on prediction market mechanics). For \fbNew, we again include the performance of OpenAI o3, the median of public surveys and superforecasters, and the best-performing LLM forecaster on \forecastbench. For \MarketLiquid, we include the performance of the prediction market price as a baseline.  These results are presented in Table~\ref{tab:headline-results} below.

\begin{table}[htbp]
\small
\centering
\begin{tabular}{lrrrr}
\toprule
\textbf{Forecaster} & \fbOldMarket & \fbOld & \fbNew & \MarketLiquid \\
\midrule
Market & 0.0965 & \texttimes & \texttimes & \textbf{0.1106}   \\
Public Survey (median per-event) & 0.1035 & 0.1451 & 0.1510 &  \texttimes  \\
Superforecasters (median per-event) & \textbf{0.0740} & 0.1110 & 0.1152 & \texttimes  \\
\forecastbench state of the art & 0.107 & 0.133 & 0.145 &  \texttimes  \\ 
OpenAI o3 & 0.1096 & 0.1221 & 0.1262 & 0.1324 \\
\systemname & 0.0753 & \textbf{0.1076} & \textbf{0.1099} & 0.1258 \\
\bottomrule
\end{tabular}
\caption{The performance of the \systemname on \forecastbench and \MarketLiquid. The \systemname is statistically indistinguishable from expert superforecasters. Uncertainty quantification is provided in Appendix \ref{app:uncertainty}.}
\label{tab:headline-results} 
\end{table}

In contrast to prior work, which shows that LLM forecasters dramatically underperform superforecasters, the \systemname is \emph{statistically indistinguishable from superforecasters} on \forecastbench. Indeed, the point estimate for our system's performance is lower (i.e., more accurate) than the superforecaster baseline. The \systemname also substantially outperforms the o3 baseline, the general public survey, and the state-of-the-art LLM-based forecaster on the \forecastbench leaderboard.

However, \forecastbench is a relatively easy forecasting benchmark. On the more challenging \MarketLiquid benchmark, the \systemname trails the performance of the market consensus. However, as we discuss in Section~\ref{sec:diversification}, the \systemname is \emph{diversifying} to this consensus, and an ensemble of the two outperforms either the market consensus or the \systemname considered separately. We now turn to arguably the most important component of a high-quality forecasting system: the search pipeline.

\section{Understanding Search in Forecasting}
\label{sec:search}

\begin{figure}
    \centering
    \includegraphics[width=0.95\linewidth]{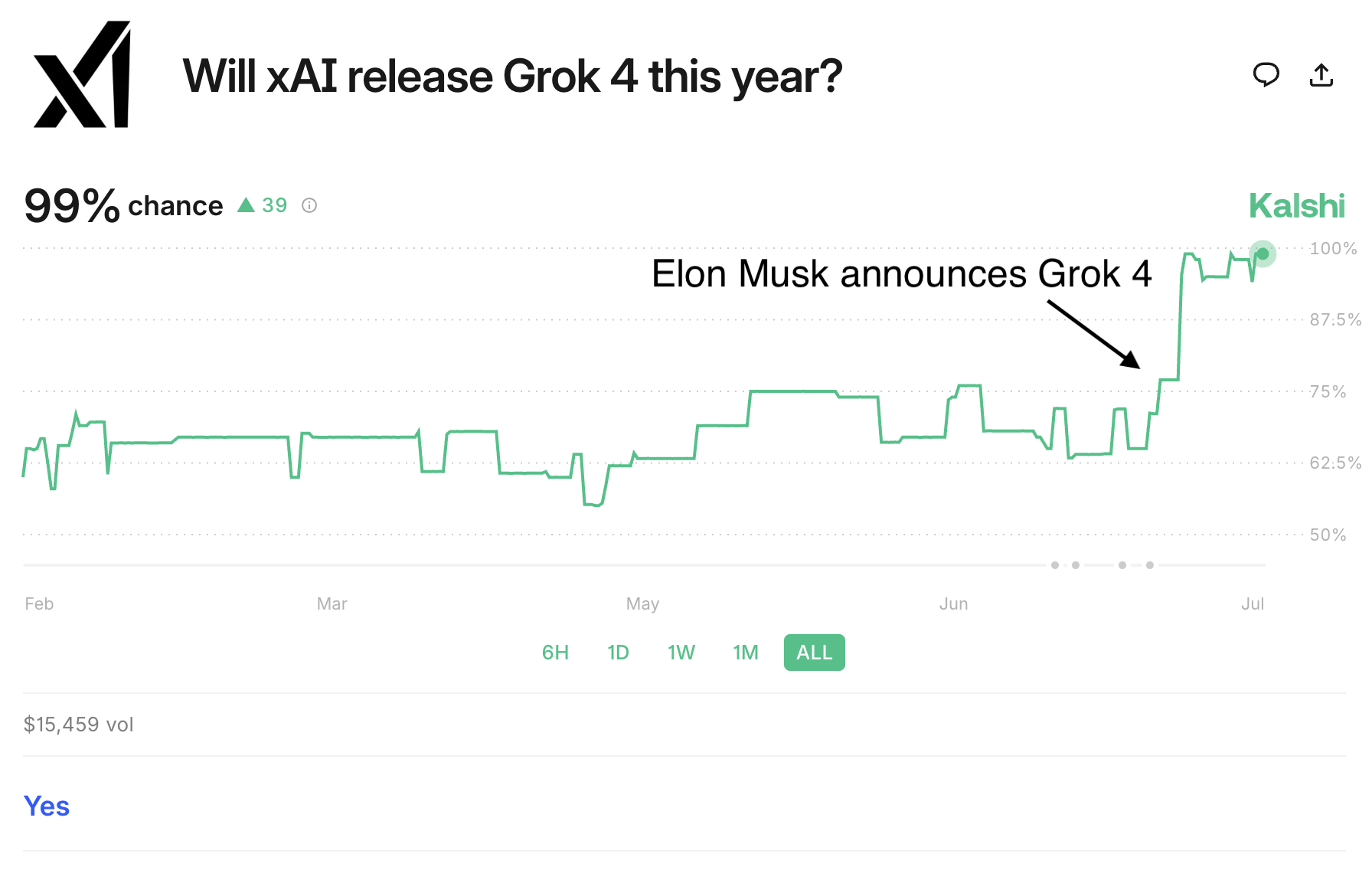}
    \caption{Example of news affecting the market price of a prediction market.}
    \label{fig:news-move}
\end{figure}

Search is a critical component of human forecasting, as timely news affects the likelihood of an event occurring. Efficient markets, by definition, aggregate salient information into market prices \citep{fama1970efficient}. We can see this effect in public prediction markets, which offer contracts that pay a fixed amount if an event occurs (and, typically, a separate contract which pays a fixed amount if it doesn't occur). This allows us to directly observe the market-implied odds of an event occurring.\footnote{We use ``market price'' and ``market-implied odds'' interchangeably, ignoring transaction costs, margin requirements, and other aspects of market structure. A trader seeking to profit from informed forecasts would of course not have this luxury.} For example, consider the question: will Grok 4 be released before the end of 2025? On June 27, 2025, Elon Musk announced that Grok 4 would be released just after July 4, 2025. This news announcement caused the market-implied odds to jump to 95-99\%, with the market resolving to yes on July 22. We show the market price movement in Figure~\ref{fig:news-move}.

However, prior work for LLM forecasters suggests that search does not affect LLM forecasting performance!  For example, \cite{karger2024forecastbench} find that the best-performing LLM-based forecaster omits search entirely, and adding a search pipeline actually \emph{reduces} forecasting accuracy. \cite{schoenegger2024wisdom} similarly do not find evidence that access to internet search improves forecasting performance. 

In this section, we show that advanced forms of search are indeed critical for performance on a wide range of markets and that the structure of prior LLM forecasting benchmarks strongly affects the ability of search to help LLM forecasters.

To do so, we must first deal with the challenge of \emph{foreknowledge bias}. All of the benchmarks described in Section \Ref{sec:main} consist entirely of events which have already occurred, which enables us to evaluate performance by comparing forecasts to ground truth outcomes. Of course, this evaluation is rendered moot if we are unable to simulate how the forecaster would have behaved in the past, by blinding it to e.g., contemporaneous news articles which describe the event in question.\footnote{\cite{sarkar2024lookahead} describe foreknowledge acquired via pre-training as ``lookahead bias.'' We use the term foreknowledge bias to include other forms of foreknowledge, including information that is exposed via search results.} We show that this is a surprisingly challenging problem and detail our mitigation strategies.

\subsection{Search Is Critical for Forecasting}
\label{sec:critical}

In this section, we quantify the impact of search on forecasting performance. As discussed above, timely information is critical for forecasting, and breaking news events can drive large movements in crowd forecasts. Furthermore, this timely information can only be incorporated in LLM-based forecasters via external tools, as the underlying models are retrained infrequently. This induces so-called ``knowledge cutoffs,'' at which point the LLM's knowledge is effectively frozen. For example, OpenAI reports a knowledge cutoff of May 31, 2024 for its flagship o3 reasoning model \citep{openai2025o3o4mini}, which means the base model has no knowledge of events occurring after this date.\footnote{The notion of a single ``knowledge cutoff" is a simplification \citep{cheng2024dated}, but nonetheless is a good heuristic for understanding where the model's general knowledge ends.}

We show that high-quality search dramatically improves forecasting performance, and that prior results to the contrary are partially explained by relatively simple, non-adaptive search pipelines. Second, we show that this effect is strongly mitigated by the practice, which is common in the literature, of directly providing prediction market prices in forecasting prompts. These market prices effectively summarize large volumes of relevant information, and can thus compensate for naive or no-search pipelines.

\textbf{Agentic search improves forecasting performance.} We first conduct a series of experiments showing that our agentic search pipeline outperforms a no-search baseline. In order to isolate the effect of search on forecaster performance, we omit the consistency analysis and Platt scaling. We further evaluate two industrial-scale search providers, which we anonymize as \ProviderA and \ProviderB. While both providers enable general-purpose semantic search over the internet, \ProviderA focuses primarily on serving natural language user queries, while \ProviderB is intended to integrate directly with agentic AI systems. We query both providers via their APIs using both ``agentic'' and ``non-agentic'' search pipelines. In the former, the LLM-based forecaster can independently define and execute a query strategy, asking follow-up questions as needed. In the latter, the researcher defines the search pipeline, first prompting the forecaster to generate a handful (here, three) search queries and then providing the results as context to subsequent LLM generations. This latter non-agentic strategy is similar to the one used in \cite{halawi2024approaching}, \cite{karger2024forecastbench}, and \cite{tao2025prophet}. 

As shown in Table~\ref{tab:retrieval_forecast_bench}, our agentic search pipeline with access to the \ProviderA API and our internal corpus of high-quality news data significantly outperforms a no-news baseline on \forecastbench (0.11435 to 0.1230).

\begin{table}[htbp]
\centering
\begin{tabular}{@{}llS[table-format=1.4]}
\toprule
\textbf{Search Method} & \textbf{Brier score}  \\
\midrule
None      & 0.1230  \\
\ProviderB (non-agentic) & 0.12168 \\
\ProviderB (agentic) & 0.11824 \\ 
\ProviderA (non-agentic)  & 0.11738  \\ 
\ProviderA (agentic) & \textbf{0.1140}  \\
\bottomrule
\end{tabular}
\caption{A comparison of different search strategies. Agentic search outperforms non-agentic search, which in turn only modestly outperforms a no-search baseline.}
\label{tab:retrieval_forecast_bench}
\end{table}

We also consistently find that, conditional on the source of news, agentic search significantly outperforms non-agentic search, and conditional on the search methodology, \ProviderA consistently outperforms \ProviderB. We conjecture that this is due to the relatively low density of relevant information in the typical web page returned by \ProviderB, while \ProviderA offers direct responses to questions that are narrowly tailored to the event in question.

We also find that the weakest search method (non-agentic queries to \ProviderB) performs comparably to simply omitting the retrieval pipeline entirely on \forecastbench. This is consistent with \cite{karger2024forecastbench} and \cite{schoenegger2024wisdom}, which find no evidence that search improves forecasting performance. However, when viewed in light of our results, this simply highlights the fact that effective search is a nontrivial problem.

\textbf{The impact of timely news is mediated by market prices.} We now show that the effects of search are strongly mitigated by providing the forecaster with access to market prices. This fact provides a partial explanation for prior work which suggests that search does not meaningfully impact forecasting performance \citep{karger2024forecastbench, schoenegger2024wisdom}.

We conduct a series of experiments in which we execute our forecaster on the cross product of: with and without search, and with and without market prices on \forecastbench. We present these results in Table \Ref{tab:market-retrieval-interaction}.

We find that simply providing the forecaster with prediction market prices improves the performance of a no-news baseline from 0.116 to 0.103 on the subset of \forecastbench derived from public prediction markets---an 11.2\% improvement. Providing the market price to our best-performing configuration improves performance from 0.085 to 0.075 on the same subset, a similar 11.8\% improvement. In these experiments, we prevent the \systemname from retrieving prediction market prices unless they are explicitly provided.\footnote{Specifically, we blacklist domains associated with the largest public prediction markets.}

\begin{table}[htbp]
\centering
\begin{tabular}{@{}llS[table-format=1.3]@{}}
\toprule
\textbf{Includes Market Price?} & \textbf{Forecasting Method} & \textbf{Brier score}  \\
\midrule
\texttimes      & LLM, no search   & 0.116 \\ 
\checkmark      & LLM, no search   & 0.103 \\ 
\checkmark      & Market price only   & 0.096 \\
\texttimes      & LLM, agentic search   & 0.085 \\
\checkmark      & LLM, agentic search   & \textbf{0.075} \\ 
\bottomrule
\end{tabular}
\caption{The interaction of market prices and search. Providing the LLM with access to market prices closes $\sim42\%$ of the gap between the no-search and agentic search baselines. This suggests that the impact of search is strongly mitigated by access to market prices, which effectively summarize large volumes of relevant information.}
\label{tab:market-retrieval-interaction}

\end{table}

We contrast these to the results of \cite{karger2024forecastbench}, who find that the best-performing LLM-based forecaster also requires access to market prices and achieves a Brier score of 0.109 on this subset.\footnote{\cite{karger2024forecastbench} report that the best overall forecaster achieves a Brier score of 0.109 on this subset. The model which performs best on this subset of questions (but worse on the full benchmark) achieves a score of 0.096 on this subset.} However, the market price alone achieves a Brier score of 0.096 on the same subset of questions; that is, the best-performing LLM-based forecaster in \cite{karger2024forecastbench} would perform better if it simply \emph{output the market price which is provided in its prompt}. The \systemname also benefits from access to the market price, but performs significantly better than using the market price alone (0.096 for the market price versus 0.075 for the \systemname).

Taken together, these results suggest caution in interpreting the performance of LLM-based forecasting systems which have direct access to prediction market prices. On the one hand, these markets can be highly efficient information aggregators, and human and artificial forecasters can both benefit from incorporating these prices into their own predictions. Prediction markets also serve as a convenient testbed for evaluating forecasting performance. On the other hand, access to market prices can mask important differences in forecasting capability, particularly when extrapolating performance to domains where no such prices are available.

\subsection{Mitigating Foreknowledge Bias in Search}
\label{sec:mitigation}

We now turn to the problem of foreknowledge bias. As discussed above, evaluating search-augmented forecasting systems presents a fundamental challenge: search providers can inadvertently leak information about events that have already occurred. In the extreme case, this bias can create the illusion of forecasting skill where none exists. 

For example, consider evaluating an LLM forecaster on the following event: ``Will Alphabet's stock price reach \$190 by December 2024?'' The resolution for this event is already known as of this writing (Alphabet stock crossed the \$190 threshold in December 2024), which allows us to evaluate forecasting skill by, e.g., blinding the LLM to events occurring after July 2024 and asking it to provide a forecast. Unfortunately, if the forecaster generates a benign-sounding search query like ``Google stock price analysis'', the search provider may return an article simply stating the factual outcome of this event. This can occur even if the search provider is instructed not to return any articles published after July 2024 --- as it turns out, determining the date when an article was published (and, perhaps, subsequently updated) is a challenging problem.

In this section, we estimate the prevalence of foreknowledge bias and bound its effect on our results. Through manual analysis, we taxonomize sources of foreknowledge bias and provide examples in Appendix~\ref{sec:sources-foreknowledge}.  We then develop an LLM-as-a-judge pipeline that reviews search provider responses and flags those that exhibit foreknowledge bias.

For each query, we collect search provider snippets and any passages cited by the forecaster, prompt the judge to indicate whether the evidence implies knowledge beyond the intended date cutoff, and record a binary flag with a brief rationale. The exact prompt and JSON schema appear in Appendix~\ref{app:foreknowledge-prompt}.

\textbf{Validation of the judge.} We manually audit $n{=}502$ search provider traces randomly sampled across both flagged and unflagged items. \autoref{tab:llm_judge_validation} summarizes performance. As intended for a high-recall detector, flags are noisy positives (low precision), while “clean” classifications are highly reliable (low false negative rate). Practically, this makes it relatively safe to trust unflagged runs, while treating raw flag counts as an overestimate of true foreknowledge bias.

\begin{table}[htbp]
\centering
\begin{tabular}{@{}lcc@{}}
\toprule
\textbf{Classification (n=502)} & \textbf{Count} & \textbf{Percentage} \\
\midrule
True Positives (correctly identified leaks) & 16 & 3.2\% \\
False Positives (flagged clean responses)   & 65 & 12.9\% \\
True Negatives (correctly identified clean) & 412 & 82.1\% \\
False Negatives (missed leaks)       & 9  & 1.8\% \\
\bottomrule
\end{tabular}
\caption{Validation of the judge on 502 randomly sampled traces. While the flags are noisy (a relatively high false positive rate), responses which are not flagged reliably exclude foreknowledge (low false negative rate).}
\label{tab:llm_judge_validation}
\end{table}

From this exercise, we estimate that $\sim 1.65\%$ of all search results include some form of foreknowledge bias. We provide additional details for this calculation in Appendix \Ref{app:blocklist}. With this estimate in hand, we now seek to bound the impact of foreknowledge bias on forecasting performance.

\textbf{Robustness checks.} We test whether residual foreknowledge bias affects our performance metrics through two conservative robustness checks:

\begin{itemize}
\item \textbf{Filtered (all flags):} Remove every judge-flagged response. Roughly 80\% of removed forecasts do not exhibit foreknowledge bias, making this test intentionally conservative.

    \item \textbf{Worst-case:} Assume a forecast of $0.50$ (corresponding to a Brier score of 0.25) for any question with $\geq$5 flags. This is again intentionally conservative, assuming no forecasting skill on any questions which appear to suffer from foreknowledge bias.
\end{itemize}

We report performance without the consistency analysis and statistical correction steps to isolate the effect of foreknowledge bias.

\begin{table}[htbp]
\centering
\begin{tabular}{@{}lcc@{}}
\toprule
\textbf{Analysis} & \textbf{Brier score} & \textbf{Change} \\
\midrule
Baseline     & 0.1159 & — \\
Filtered                   & 0.1161 & +0.17\% \\
Worst-case                 & 0.1166 & +0.55\% \\
\bottomrule
\end{tabular}
\caption{Robustness checks to evaluate foreknowledge bias on \forecastbench. ``Filtered'' computes the Brier score after removing all questions flagged by the judge. ``Worst-case'' assumes a forecast of 0.5 for all flagged questions. Both perform comparably to a no-mitigation baseline.}
\label{tab:leakage_impact}
\end{table}

As Table \Ref{tab:leakage_impact} demonstrates, both robustness checks yield a Brier score that is within $0.001$ of the baseline value ($\le0.6\%$ on a relative basis). Thus, we conclude that foreknowledge bias does not appear to materially affect performance. 

\subsection{LLM Forecasts Are Diversifying to Market Prices}
\label{sec:diversification}

We now investigate the complementarity between LLM forecasts and market prices. In Section~\ref{sec:main} we saw that our forecaster outperforms the market consensus on \forecastbench, but modestly underperforms on the stronger \MarketLiquid benchmark. It is natural to then ask: in cases where the forecaster outperforms the market, does the market price provide any \emph{additional} predictive signal? And conversely, in settings where the forecaster underperforms the market consensus, can we disregard its forecasts entirely? 

We investigate these questions by estimating a bivariate simplex-constrained regression of the binary resolution indicator on two covariates: the market price and the LLM forecast. This regression learns a convex combination of the two covariates that minimizes the Brier score (i.e., the standard $\ell_2$ loss). Our goal is not to produce the best possible ensemble forecast, but rather to derive interpretable coefficients that serve as proxies for the relative information each forecast provides about the event of interest. We report point estimates for these coefficients along with confidence intervals computed from 1000 bootstrap samples. To contextualize these results, we include the Brier score achieved by a leave-one-out version of this simplex regression procedure, providing an out-of-sample estimate of the ensemble's performance.

\begin{table}[htbp]
\centering
\begin{tabular}{@{}l|cc|cc|c@{}}
\toprule
 & \multicolumn{2}{c|}{\textbf{AIA Forecaster}} & \multicolumn{2}{c|}{\textbf{Market}} & \textbf{Ensemble} \\
\textbf{Benchmark} & Brier score & Coefficient & Brier score & Coefficient & Brier score \\
\cmidrule(lr){1-6} 
\fbOldMarket & \textbf{0.075} &  0.87 [0.42, 1.00] & 0.097 & 0.14 [0.00, 0.58] & 0.079 \\
\MarketLiquid      &  0.126  &  0.33 [0.12, 0.47] & 0.111 & 0.67 [0.53, 0.80] & \textbf{0.106} \\
\bottomrule
\end{tabular}
\caption{The performance of a simplex-constrained regression ensemble which combines LLM forecasts with the market price. Regression coefficients are reported with $95\%$ bootstrap confidence intervals over 1000 samples. Ensemble performance is reported as a leave-one-out estimate. Coefficients may not sum to $1$ due to rounding.}
\label{tab:ensemble}
\end{table}

As Table~\ref{tab:ensemble} demonstrates, the \systemname is assigned the majority of the ensemble weight on the \fbOldMarket benchmark, and the leave-one-out ensemble performs worse than the \systemname alone. Thus, the market price does \emph{not} provide signal that is additive to the forecasts produced by our system. However, on \MarketLiquid, where the market price outperforms the \systemname, the converse is not true. Instead, the ensemble assigns roughly a third of its weight to the LLM forecasts, and the leave-one-out ensemble outperforms both the LLM forecasts and the market price. This suggests that even in challenging settings where LLMs fail to outperform the market consensus outright, they can still provide valuable and diversifying information to decision makers.

\section{Search Is Critical for Live Markets}
\label{sec:live_markets}

We now turn to evaluating the \systemname on prediction markets which have not yet resolved (``live markets'') at the time of forecaster execution. This kind of evaluation precludes any form of foreknowledge bias, as we are asking the model to predict events which have not yet occurred. 

To do this, we collected a dataset of live markets on a popular prediction market platform, filtered by relevance and liquidity. Our filtering process for the live markets was the same process used to generate the static \MarketLiquid dataset (except filtered to markets that are currently open). We conducted this process nightly between August 15, 2025 and August 21, 2025, which yielded approximately 1500 questions per night. We sampled to 250 questions at random every night to generate the evaluation set. These questions changed on a nightly basis as new markets opened and existing markets were resolved.

We conduct two sets of experiments. For both experiments, we executed the \systemname with and without search enabled. In both conditions, we omitted the market prices to isolate the effect of search on forecaster performance.\footnote{As in Section \Ref{sec:critical}, we exclude market prices from the prompt and prevent the forecaster from accessing domains associated with the most popular prediction market platforms.}

\begin{table}[htbp]
\centering
\begin{tabular}{@{}lS[table-format=1.4]S[table-format=1.4]S[table-format=1.4]@{}}
\toprule
\textbf{Market Type} & \textbf{With Search} & \textbf{Without Search} & \textbf{Market Baseline} \\
& {(Brier score)} & {(Brier score)} & {(Brier score)} \\
\midrule
Closed markets (n=64) & \textbf{0.1002} & 0.3609 & 0.1111 \\
Open markets (n=1750) & \textbf{0.0522} & 0.0910 & {---} \\
\bottomrule
\end{tabular}
\caption{Real-time forecasting performance with and without search. In both cases the system is prevented from accessing market prices. In closed markets, forecasts are evaluated against ground truth outcomes. In open markets, forecasts are scored against market prices. The performance of the market consensus on closed markets is included for reference. \systemname improves on the no-search baseline by 3.6$\times$ in closed markets and is more concordant with market prices in open markets.}
\label{tab:live_markets_results_updated}
\end{table}

In the first experiment, we evaluate the Brier score on the 64 markets that closed by August 26, 2025. As shown in Table~\ref{tab:live_markets_results_updated}, with search, the \systemname achieved a Brier score of 0.1002. Without search, the \systemname achieved a Brier score of 0.3609, a difference of 3.6$\times$. As we can see, search is critical for forecasting; in fact, excluding search yields performance which is worse than uniformly predicting $50\%$ (which mechanically achieves a Brier score of $0.25$).

We also compute the Brier score achieved by simply using the market price as a baseline (``consensus'') forecast. We see that, with search, the \systemname slightly outperforms the market consensus, although with only 64 markets we are unable to statistically distinguish these two results.

In the second experiment, we evaluate the Brier score \emph{with respect to the market price}. Unlike the Brier score for resolved markets, a perfectly skilled forecaster would score substantially away from 0. Nonetheless, market prices are a strong baseline forecast \citep{Surowiecki2004, bassambooWisdom2015, zhi2019wisdom, lichtendahl2013wisdom}, so significant deviations from a market price baseline are unlikely for a skilled forecaster. Importantly, this allows us to evaluate the \systemname over the much larger (1750) set of markets which are still open.

As shown in Table~\ref{tab:live_markets_results_updated}, compared to market prices, \systemname achieves a Brier score of 0.0522 with search enabled. Without search, our forecaster achieves a Brier score of 0.0910. Because we expect market prices to be informative, being close to the market price likely indicates better performance. Thus, we can again see that search is critical for generating informative forecasts.

These experiments on live markets are necessarily narrower in scope than the larger-scale retrospective evaluations we conduct in Section \Ref{sec:main}. Nonetheless, we again see evidence which corroborates our main findings, and further suggests that foreknowledge bias is not meaningfully impacting our results.

\section{Ensembling, Synthesis, and Statistical Corrections}
\label{sec:synthesis}

In this section, we explore various approaches to ensembling multiple forecasts into a stronger aggregate forecast. We find that some form of ensembling is critical and dramatically outperforms naive baselines. In Section~\ref{sec:simple_averaging}, we motivate the need for ensembling forecasts, and find that simple averaging of independent forecasts (and minor variations thereof) dramatically outperforms a naive baseline which eschews any form of aggregation. In Section~\ref{sec:supervisors}, we find that an additional ``supervisor'' agent can substantially improve on the performance of simple mean aggregation by actively resolving disagreements between individual forecasts. Finally, in Section~\ref{sec:stats}, we find that the additional application of Platt scaling improves our performance to match that of superforecasters.

\subsection{Simple Averaging Is a Strong Baseline}
\label{sec:simple_averaging}

We now explore the performance of simple aggregation methods. First, to motivate the need for aggregating multiple forecasts, consider the following simple model. Let the random variables $Q \in \mathcal{Q}, P \in [0, 1], O \in \{0, 1\}$ denote questions, forecasts and outcomes, respectively. Let $(Q, P, O)$ be a random vector with some well-defined probability measure. Let $\ell(P, O) = (O - P)^2$ be the Brier score. Under this model, fixing a question $Q = q$ induces some conditional distribution over the question-specific Brier score $\ell$. Because $\ell$ is strongly convex, the following is an immediate consequence of Jensen's inequality:

\begin{align}
    \ell\left(\mathbb{E}[P \mid Q = q], \mathbb{E}[O \mid Q = q] \right) < \mathbb{E}\left[\ell\left(P, O\right) \mid Q = q\right] \qquad \forall q.
\end{align}

That is, the Brier score for the average forecast and outcome is a strict improvement on the Brier score averaged over possible forecasts and outcomes.\footnote{\cite{bagging} uses this fact to motivate ensembling in the classical statistical learning paradigm.} For example, given the choice between (1) predicting exactly $50\%$ or (2) predicting $100\%$ and $0\%$ with equal probability, the former is preferable \emph{regardless of the true outcome}. This suggests that averaging many independent forecasts for a given question will tend to improve on sampling a single forecast. We confirm this empirically in Figure \Ref{fig:ensemble} below, which demonstrates sharply decreasing Brier scores on \MarketLiquid as we increase the number of independent forecasts from 1 to 5, and an additional modest improvement as we further increase the number to 15 forecasts.\footnote{To maintain computational tractability, Figure \Ref{fig:ensemble} presents the results of bootstrap resampling a fixed number of forecasts from superpopulation of 50 forecasts per question rather than repeatedly re-evaluating the benchmark at each point on the x-axis. We use the larger \MarketLiquid benchmark to increase statistical power.}

\begin{figure}[htbp]
    \centering
    \includegraphics[width=0.75\linewidth]{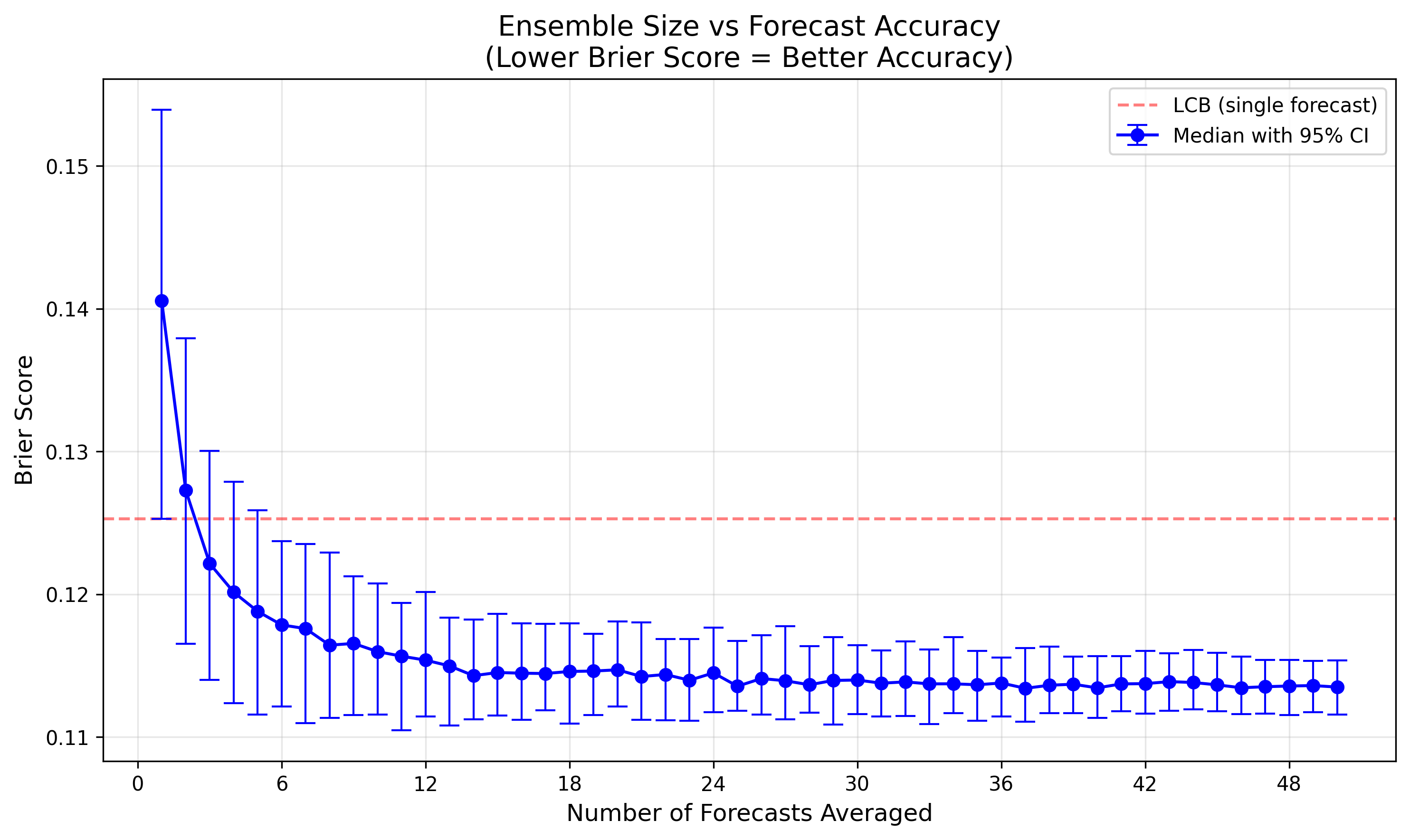}
    \caption{The Brier score induced at various ensemble sizes. Point estimates and $95\%$ confidence intervals are generated via bootstrap resampling from a set of 50 forecasts per question. The dashed line indicates the lower confidence bound for a single forecast.}
    \label{fig:ensemble}
\end{figure}

Figure \Ref{fig:ensemble} also suggests diminishing returns beyond a handful of independent forecasts; we adopt $10$ forecasts per question as a conservative standard in this work. This result also illustrates that sampling only a single forecast for each question yields highly variable performance---even when averaged over the 1610 questions in this evaluation set---due to the inherent randomness in LLM outputs. This variability is itself an undesirable property for a forecaster, and suggests that some form of ensembling is critical for internal validity. This kind of ensembling has thus far not been common in the LLM forecasting literature\footnote{One exception is \cite{schoenegger2024wisdom}, though this work is motivated by the hypothesis that ensembling \emph{different} LLM-based forecasters will improve performance.}, and suggests that the common practice of quantifying uncertainty \emph{only over forecasting questions themselves}---rather than also considering uncertainty over LLM outputs---dramatically understates the true variability of LLM forecasting performance.

In Table \Ref{tab:ensembling}, we further demonstrate that minor variations of this approach, like taking the median of 10 forecasts, or taking the trimmed mean (by discarding the smallest and largest of the 10 forecasts), perform comparably to simple averaging. We include a reference baseline which samples a single individual forecast for each question uniformly at random.

\begin{table}[htbp]
\centering
\begin{tabular}{@{}lS[table-format=2.2]S[table-format=2.2]@{}}
\toprule
\textbf{Synthesis Method} & \textbf{Brier score} \\
\midrule
Simple Mean & 0.1140 \\
Trimmed Mean & 0.1142 \\
Median & 0.1138 \\
Single Forecast &  0.1182 \\ 
\bottomrule
\end{tabular}
\caption{Evaluating different approaches to ensembling multiple forecasts on \fbOld. The mean, trimmed mean and median aggregation over 10 forecasts all perform substantially better than a single randomly chosen forecast.}
\label{tab:ensembling}
\end{table}

The results above suggest that averaging multiple independent forecasts is critical for forecasting performance and cannot be meaningfully improved by minor variations thereof.

\subsection{Supervisor Agents Resolve Disagreements}
\label{sec:supervisors}

We now turn to more sophisticated ensembling methods, which utilize an LLM-based supervisor to synthesize multiple independent forecasts. This supervisor can be viewed as a kind of generative verifier or LLM-as-a-judge \citep{li2024judge, zhang2024verifiers} which aggregates the 10 forecasts generated by each agent into a final probability. 

We consider three different supervisor agents, as well as the same no aggregation baseline described in Section~\ref{sec:simple_averaging}. The first, which we call the ``best of $k$'' supervisor, is provided with 10 forecasts and asked to reason over them to output the best of these forecasts. Importantly, the supervisor is explicitly constrained to output exactly one of the provided forecasts. The second, which we call the ``non-agentic supervisor,'' uses the method proposed in \cite{halawi2024approaching} to ensemble multiple forecasts into a final forecast. Unlike the best of $k$ method, this method is not constrained to output one of the input forecasts. Instead, the model simply uses these as inputs to reason toward a stronger forecast. Finally, we propose the ``agentic supervisor,'' which proceeds in three steps. First, the supervisor model identifies disagreements among the 10 input forecasts. Second, the supervisor generates and executes a series of search queries intended to resolve these disagreements (e.g., to resolve disagreement over the accurate base rate for an event). Finally, the supervisor generates an updated forecast along with its confidence that the forecast was updated in the correct direction (``high,'' ``medium'' or ``low''). High confidence updates are used in place of the simple mean, while medium or low confidence updates are discarded.

\begin{table}[htbp]
\centering
\begin{tabular}{lrrrr}
\toprule
\textbf{Synthesis Method} & \textbf{Brier score} & \textbf{Top@3}  & \textbf{Worst@3} & \textbf{Outperform (\%)} \\
\midrule
Agentic Supervisor & \textbf{0.1125} & \textbf{33.9\%} & \textbf{4.6\%} & \textbf{3.8\%} \\
Non-agentic Supervisor & 0.1168 & 7.8\% & 6.2\% & 0.6\% \\
Best of $k$ (LLM) & 0.1191 &  29.7\% & 7.2\% & 0.0\% \\
None & 0.1199 \\ % (choose single random forecast)
\bottomrule
\end{tabular}
\caption{Comparison of different synthesis strategies on \fbOld. \textbf{Top@3}: fraction of aggregated forecasts which are at least as accurate as the three best individual forecasts; \textbf{Worst@3}: fraction of aggregated forecasts which are no more accurate than the three worst individual forecasts; \textbf{Outperform (\%)}: percentage of aggregate forecasts beating all individual forecasts. Agentic supervision yields the best overall performance.}
\label{tab:supervisor_agent}
\end{table}

As shown in Table~\ref{tab:supervisor_agent}, the agentic supervisor achieves the best performance among the synthesis methods. The best of $k$ approach, which selects a single forecast from the ensemble to serve as the final prediction, performs poorly. It never selects a forecast that is better than any of the candidate forecasts by design, and it happens to pick among the worst of the candidate forecasts with a striking 7.2\% frequency. The non-agentic supervisor, in which a language model performs the aggregation, shows modest improvement over the best of $k$ method. However, this method produces quality forecasts (top@3 of 7.8\%) relatively rarely. In contrast, the agentic supervisor substantially outperforms both alternatives. It actively reconciles inconsistencies between individual predictions and performs additional news queries to refine forecast rationales. This synthesis process enables it to achieve the lowest Brier score of 0.1125, highest top@3 accuracy, highest outperformance rate, and lowest worst@3 rate, demonstrating the value of additional search and agentic reasoning when aggregating forecasts.

\subsection{Statistical Corrections and Where They Help}
\label{sec:stats}

\begin{figure}[htpb!]
    \centering
    \includegraphics[width=0.9\linewidth]{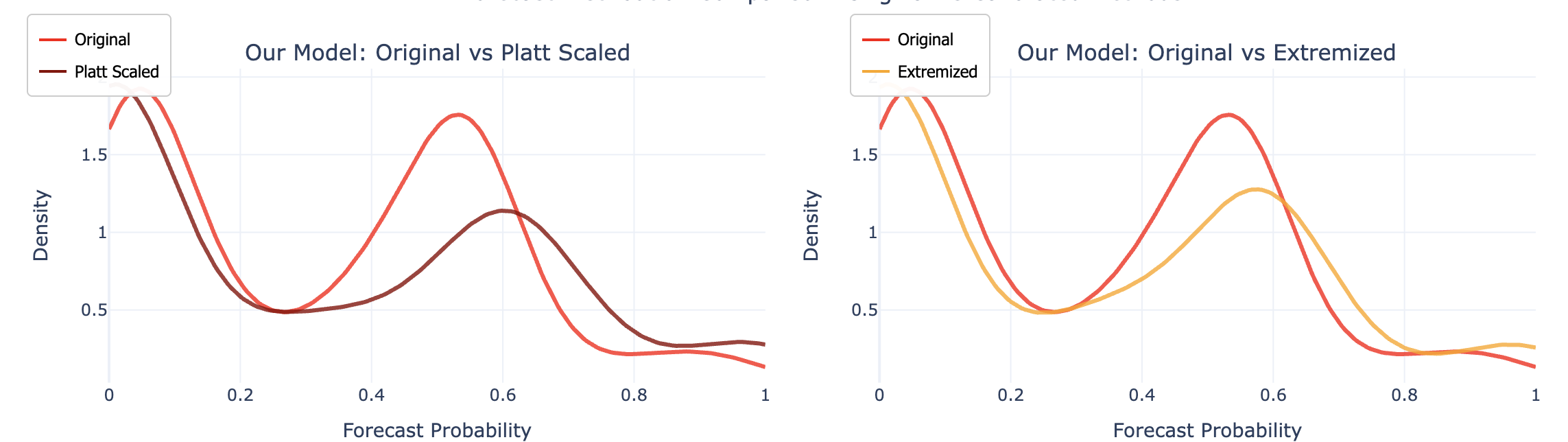}
    \caption{Scaling and extremization shifts the mass of probabilities toward the extremities, especially in the center mass which produces larger drops in Brier scores.}
    \label{fig:forecast_shift_extreme}
\end{figure}

Statistical corrections are commonly applied to forecasts to correct for distortions in the forecasted probabilities \citep{good-probs, tetlock-extreme-paper}. In the binary case, \citeauthor{good-probs} note that forecasts often result in probabilities that are clustered away from zero or one, which we observe in our data (Figure \ref{fig:forecast_shift_extreme}).

We consider statistical correction functions $f_{\theta}(p): [0,1] \to [0,1]$ \citep{random-expert}, which may either have learnable parameters \citep{good-probs} or are deterministic functions \citep{logit-combination}. 

\begin{table}[htbp]
\centering
\begin{tabular}{lrrr}
\toprule
\textbf{Correction Method} & \textbf{Fixed Parameter} & \textbf{ID Training} & \textbf{OOD Training} \\
\midrule
Platt Scaling & \textbf{0.1076} & \textbf{0.1071} & \textbf{0.1104} \\
Log Odds Extremization & 0.1085 & \texttimes & \texttimes \\
Isotonic Regression & \texttimes & 0.1097 & 0.1134 \\
OLS & \texttimes & 0.1119 & 0.1125 \\
None & 0.1140 &  \texttimes & \texttimes \\
\bottomrule
\end{tabular}
\caption{Evaluating different statistical correction approaches on \fbOld. The fixed parameter setting sets the scaling parameter to $\sqrt{3}$, as proposed by \cite{random-expert}. In distribution (ID) training reports leave-one-out performance with learned scaling parameters. Out of distribution (OOD) training fits the scaling parameters on the benchmark released by \cite{halawi2024approaching}. Platt scaling yields the best performance across these methods.}
\label{tab:stat_corrections}
\end{table}

Table \ref{tab:stat_corrections} shows that the most beneficial statistical correction methods are Platt scaling \citep{good-probs} and extremization \citep{random-expert}. Log odds extremization aggregates a set of $k$ forecasts by taking a scaled mean of the log odds of each forecast. The scaling parameter can be learned from data or set to a fixed value a priori. Log odds extremization follows the following formula \citep{tetlock-extreme-paper}:
\begin{equation}
\label{eqn:extremization}
\log \frac{\hat{p}}{1-\hat{p}} = \frac{d}{n} \sum_{i=1}^n \log \frac{p_i}{1-p_i},
\end{equation}

where the aggregate estimate $\hat{p}$ can then be recovered using the sigmoid function. We demonstrate that Equation \ref{eqn:extremization} is mathematically identical to applying Platt scaling to the geometric mean of forecasts with a Platt coefficient of $d$ in Appendix~\ref{app:platt-extreme}.

\cite{random-expert} propose a parameter value of $d = \sqrt{3}$. We adopt this convention in our work to avoid the risk of overfitting to the benchmarks we consider. Using Platt scaling on the arithmetic mean of $k$ forecasts results in the lowest Brier score, as shown in Table \ref{tab:stat_corrections}.

\begin{figure}[htpb!]
    \centering
    \includegraphics[width=0.75\linewidth]{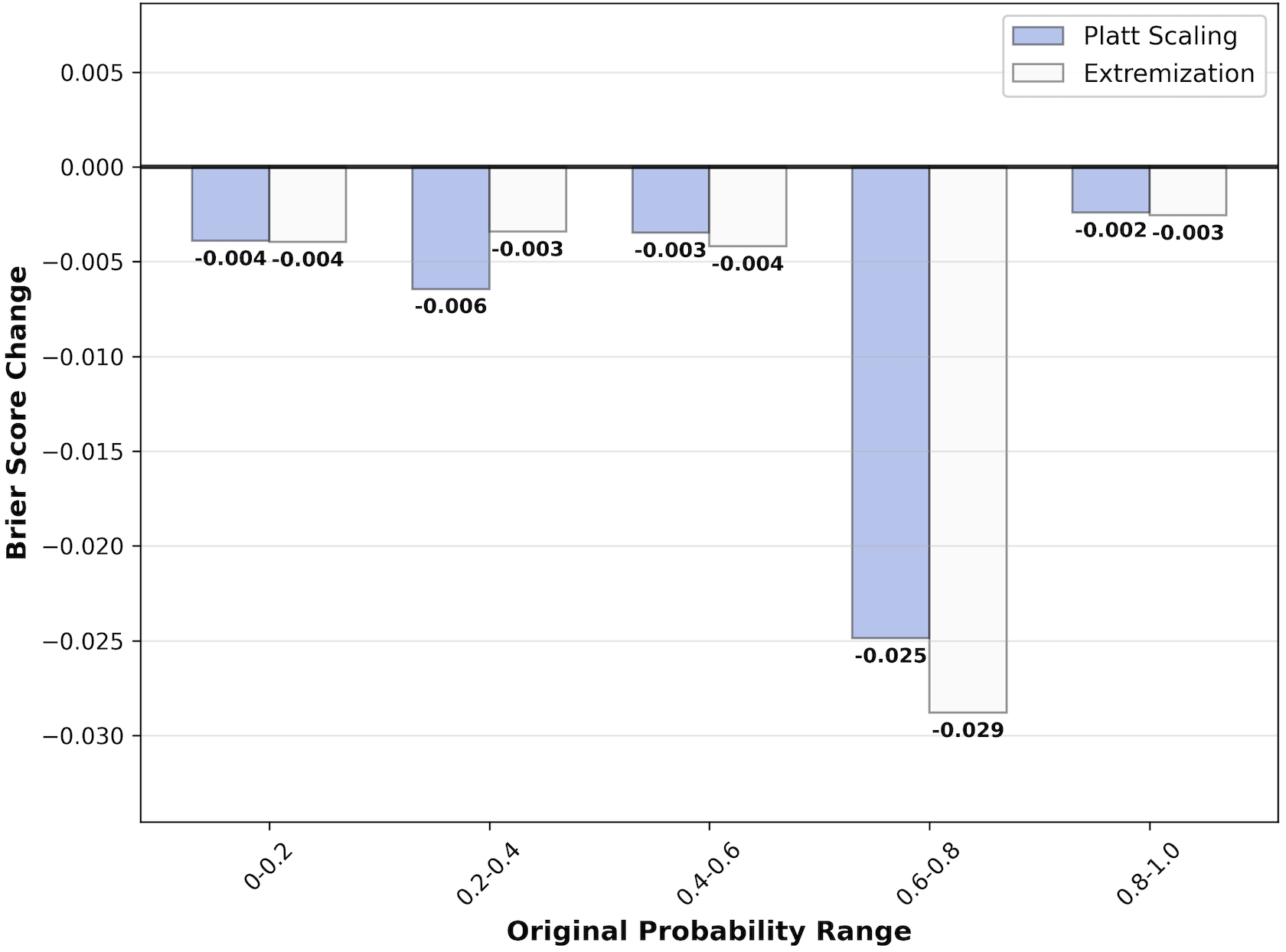}
    \caption{Across probability bins, the largest drops in Brier scores due to correction comes from the 0.6-0.8 forecast bin, followed by the 0.2-0.4 forecast bin.}
    \label{fig:improvement_bins}
\end{figure}

\textbf{Mechanisms of Improvement.} As Table \Ref{tab:stat_corrections} shows, both Platt scaling and extremization significantly improve forecast accuracy, while isotonic regression and OLS are less competitive. Both Platt scaling and extremization effectively push probabilities toward their respective bounds (0 or 1). This suggests that initial forecast quality is critical to these kinds of statistical corrections, insofar as forecasts on the wrong side of 0.5 would be adjusted in the wrong direction. We also see that the improvement gained by original forecasts that are already confident and correct are (mechanically) limited. For example, an increase in the forecast from $0.995$ to $0.999$ yields a corresponding change in the Brier score of $2.4\times 10^{-5}$. Meanwhile a change in the forecast from $0.497$ to $0.381$ yields a corresponding change in the Brier score of $-0.102$. That is, Platt scaling and extremization push forecasts towards the extremes. Figure \ref{fig:forecast_shift_extreme} illustrates this redistribution effect, where mass shifts from the 0.5-0.6 range to the 0.6-0.75 range for events that ultimately resolve to 1.0. 

This push toward the tails is what generates the substantial Brier score improvements when less confident but correct predictions become more confident, which we can also observe in Figure \ref{fig:improvement_bins}\footnote{Near $p \approx 0.5$, Platt scaling and extremization are approximately the identity function. In such scenarios, the brier score difference is minimally impacted as this effect dominates the quadratic Brier score improvement.}, where the largest Brier score decreases come from original forecasts $p \in [0.2, 0.4) \cup [0.6, 0.8)$. 

Although extremization and Platt scaling are both effective and share mathematical similarities, supervisor agents offer significant improvements to forecast quality, as discussed in Section~\ref{sec:supervisors}. The extremization approach detailed in Equation~\ref{eqn:extremization} requires aggregating $k$ forecasts, which prevents us from leveraging the supervisor's contributions. Therefore, we adopt Platt scaling as our correction of choice: it preserves the advantages of extremization while enabling the supervisor resolution step, ultimately yielding more accurate forecasts as demonstrated in Table~\ref{tab:stat_corrections}.

\begin{figure}[htpb!]
    \centering
    \includegraphics[width=0.75\linewidth]{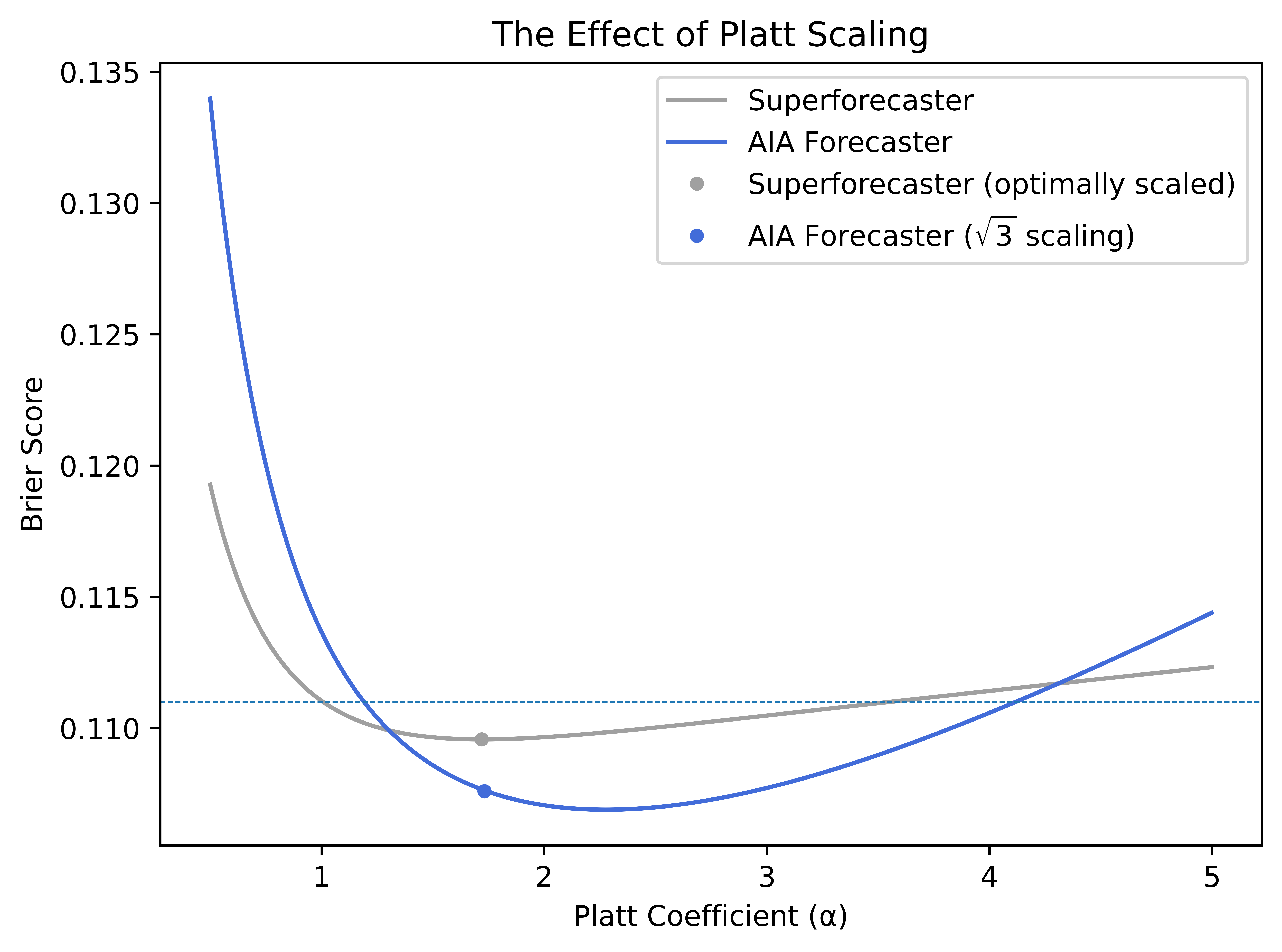}
    \caption{\systemname outperforms superforecasters across a range of Platt coefficients, including when the coefficient is optimized for superforecaster performance ($\alpha = 1.72$, gray dot). The \systemname uses a coefficient of $\alpha = \sqrt{3} \approx 1.73$, as suggested by \citet{random-expert} (blue dot).}
    \label{fig:platt-scaling}
\end{figure}

\textbf{Comparison with Superforecasters.} Finally, because statistical corrections are agnostic to the technique used to generate the forecasts, they can also be applied to the superforecaster predictions. Indeed, applying a statistical correction to the superforecaster predictions is arguably required to produce a fair comparison between human and AI forecasts \citep{nikos2024contra}.

In Figure \Ref{fig:platt-scaling}, we plot the performance of both superforecasters and the \systemname across a range of Platt coefficients. As Figure \Ref{fig:platt-scaling} demonstrates, the \systemname outperforms the superforecaster baseline across a wide range of scaling parameters. Furthermore, scaling the \systemname with a Platt coefficient of $\sqrt{3} \approx 1.73$, as suggested by \citet{random-expert}, outperforms superforecasters even when the scaling parameter is chosen to optimize their performance (a coefficient of $1.72$). Optimizing the scaling coefficient for the \systemname yields a further modest improvement (at a coefficient of $2.27$), though, as described above, we do not use this learned parameter to mitigate the risk of overfitting to a particular benchmark.

\section{Discussion}
\label{sec:discussion}

This work describes a practical recipe for LLM-based judgmental forecasting: (i) prioritize evidence quality via agentic, adaptive search; (ii) reduce variance by ensembling independent runs; (iii) use a supervisor agent to actively reconcile disparate forecasts; and (iv) apply statistical calibration to counteract the behavioral biases of LLMs. Together, these components yield performance that is statistically indistinguishable from human superforecasters on \forecastbench, and diversifying to (albeit less accurate than) market consensus in liquid prediction markets.

Methodologically, we find that ensembling is not optional. Single-run performance is both noisier \emph{and worse in expectation} than even simple averaging across a small number of independent generations. This also implies that the practice---which has thus far been common in the LLM forecasting literature---of neglecting to quantify uncertainty of LLM outputs themselves may threaten both internal and external validity of the claimed results. Second, generative verification methods are not necessarily superior to simple averaging. In fact, naive implementations of this approach can perform quite poorly. However, an agentic supervisor that can actively resolve disagreements and seek clarifications yields meaningful gains in forecasting performance. Third, statistical corrections provide a lightweight mechanism for aligning forecasts with scoring rules. However, post hoc calibration only helps when initial judgments are already on the correct side of 0.5, reinforcing that upstream evidence and reasoning are the primary mechanisms for improving forecast quality.

\textbf{Limitations, risks and future work.} A primary theme of this work is the difficulty of evaluating LLM-based forecasters. Perhaps the biggest challenge is the narrow temporal window in which this evaluation can be performed. On the one hand, events occurring before the model's knowledge cutoff are likely to have been memorized by the LLM \citep{sarkar2024lookahead}, making retrospective analysis difficult. On the other hand, evaluating the LLM on future events is inefficient---we need to wait for the events to occur!---which precludes rapid iteration. This challenge is compounded by the more general issue of foreknowledge bias in search results \citep{paleka2025pitfalls}. LLMs are also known to be prone to hallucination \citep{huang2025hallucination} and do not always reason in an internally consistent manner \citep{valmeekam2023planning, vafa2024world}. This makes it difficult to extrapolate from performance on one task to other (seemingly) related domains, and we caution that our work does not in any way obviate the need for human oversight in high-stakes decision tasks. 

Finally, our work focuses on forecasting \emph{unconditional}, \emph{binary} events, of the form ``will X happen by Y date?'' There are many other types of forecasting tasks, including k-class forecasting (e.g., ``which equity sector will perform the best over the next year?''), point estimate forecasting (e.g., ``what will the next inflation print be?'') and conditional forecasting (e.g., ``if the US enters a recession, what will Donald Trump's approval rating be?''). All of these alternative forms of forecasting are amenable, in principle, to LLM-based forecasting, but will require significant additional work to incorporate into our forecasting system. 

Despite these limitations, our work demonstrates the significant promise of LLM-based forecasting. We achieve expert-level performance on the \forecastbench benchmark, and resolve a number of open questions in the literature. We also develop a set of practical, easily portable standards for LLM-based forecasting, including the use of agentic search, ensembling and statistical calibration techniques. Taken together, our findings suggest that AI forecasters may soon achieve superhuman performance.

\newpage

\bibliography{references}

\newpage

\appendix
\addcontentsline{toc}{section}{Appendix}                                                                                                              
 \refstepcounter{section}  % Increment counter to A (1st appendix)
\section*{A. Uncertainty Quantification for Primary Results}
\addcontentsline{toc}{subsection}{A. Uncertainty Quantification for Primary Results}
\label{app:uncertainty}
\setcounter{subsection}{0}  % Reset subsection counter    

In this section we quantify uncertainty over the results presented in Section~\ref{sec:main}. To do this, we follow \cite{karger2024forecastbench} in computing pairwise p-values via bootstrap resampling. In particular, we first compute the mean difference in Brier scores at the individual question level, which will serve as our statistic of interest. This is equivalent to simply computing the difference in overall Brier score that two different methods achieve on a given benchmark. We then approximate the null distribution of this statistic by (1) recentering these question-level differences at zero, (2) drawing 10000 bootstrap samples from these recentered differences, and (3) computing the mean difference in Brier score for each of these samples. This procedure will, for example, have power to detect cases where one forecaster performs slightly better than another \emph{uniformly} over all questions, even if the per-question performance gap is small (e.g., the better forecaster always provides a forecast that is $1\%$ closer to the true outcome). In contrast, computing separate confidence intervals for each Brier score, without conditioning on the questions, will struggle to detect this kind of performance gap. Importantly, this procedure does not account for uncertainty over the LLM outputs themselves. We address this separately Section~\ref{sec:synthesis}. We present these results in Tables~\ref{tab:fb-original-uncertainty-market}, \ref{tab:fb-original-uncertainty}, \ref{tab:fb-updated-uncertainty} and \ref{tab:market-uncertainty} below.

\begin{table}[htbp]
\centering
\begin{tabular}{@{}lrrr@{}}
\toprule
\textbf{Forecaster} & \textbf{Brier score} &  \textbf{Win Rate vs Best} & \textbf{Pairwise p-value}\\
\midrule 
Market & 0.0956 & 0.4212 & 0.0687 \\
Public Survey (median per-event) & 0.1035 & 0.2236 & <0.001 \\
\cite{karger2024forecastbench} state of the art & 0.107 & - & - \\
OpenAI o3 & 0.1096 & 0.6447 & 0.0305 \\
\textbf{\systemname} & 0.0753 & 0.3026 & 0.4328 \\
Superforecasters (median per-event) & \textbf{0.0740} & - & - \\

\bottomrule
\end{tabular}
\caption{Uncertainty quantification for results on \fbOldMarket in Table \Ref{tab:headline-results}. \cite{karger2024forecastbench} do not publish individual LLM forecasts, precluding a direct pairwise comparison. }
\label{tab:fb-original-uncertainty-market} 
\end{table}

\begin{table}[htbp]
\centering
\begin{tabular}{@{}lrrr@{}}
\toprule
\textbf{Forecaster} & \textbf{Brier score} &  \textbf{Win Rate vs Best} & \textbf{Pairwise p-value}\\
\midrule 
Public Survey (median per-event) & 0.1451 & 0.2149 & <0.001 \\
Superforecasters (median per-event) & 0.1110 & 0.3474 & 0.1522 \\
\cite{karger2024forecastbench} state of the art & 0.133 & - & - \\
OpenAI o3 & 0.1221 & 0.4839 & 0.0025 \\
\textbf{\systemname} & \textbf{0.1076} & - & - \\
\bottomrule
\end{tabular}
\caption{Uncertainty quantification for results on \fbOld in Table \Ref{tab:headline-results}. \cite{karger2024forecastbench} do not publish individual LLM forecasts, precluding a direct pairwise comparison. }
\label{tab:fb-original-uncertainty} 
\end{table}

\begin{table}[htbp]
\centering
\begin{tabular}{@{}lrrr@{}}
\toprule
\textbf{Forecaster} & \textbf{Brier score} &  \textbf{Win Rate vs Best} & \textbf{Pairwise p-value}\\
\midrule

Public Survey (median per-event) & 0.1510 & 0.2209 & <0.001\\ 
Superforecasters (median per-event) & 0.1152 & 0.5066 & 0.0946 \\
\cite{karger2024forecastbench} state of the art & 0.145 & - & - \\
OpenAI o3 & 0.1262 & 0.4618 & <0.001 \\
\textbf{\systemname} & \textbf{0.1099} & - & - \\
\bottomrule
\end{tabular}
\caption{Uncertainty quantification for results on \fbNew in Table \Ref{tab:headline-results}. \cite{karger2024forecastbench} do not publish individual LLM forecasts, precluding a direct pairwise comparison.}
\label{tab:fb-updated-uncertainty} 
\end{table}

\begin{table}[htbp]
\centering
\begin{tabular}{@{}lrrr@{}}
\toprule
\textbf{Forecaster} & \textbf{Brier score} &  \textbf{Win Rate vs Best} & \textbf{Pairwise p-value}\\
\midrule 
OpenAI o3 & 0.1324 & 0.7202 & 0.0016 \\
AIA Forecaster & 0.1258 & 0.7664 & 0.0567 \\
\textbf{Market} & 0.1106 & - & - \\
\bottomrule
\end{tabular}
\caption{Uncertainty quantification for results on \MarketLiquid in Table \Ref{tab:headline-results}.}
\label{tab:market-uncertainty} 
\end{table}

 \refstepcounter{section}  
\section*{B. Evaluating Different Base Models}
\addcontentsline{toc}{subsection}{B. Evaluating Different Base Models}
\label{app:base_models}
\setcounter{subsection}{0}  % Reset subsection counter                     

In this section we record the performance of different base models on the \MarketLiquid. We choose this benchmark because it includes very recent events, all of which can be evaluated after the knowledge cutoff of the models we consider.\footnote{Claude Sonnet 4 has the latest knowledge cutoff, which is March 2025. However, its ``reliable'' knowledge cutoff is January 2025. The earliest questions in the \MarketLiquid benchmark are asked ``as of'' 01/01/2025. This complicates the interpretation of the Sonnet 4 results, which likely include some form of lookahead bias \citep{sarkar2024lookahead}.}  We exclude the supervisor agent (see Section~\ref{sec:supervisors}) to isolate the quality of the baseline reasoning, and otherwise use the pipeline described in Section~\ref{sec:methods}.

\begin{table}[htbp]
\centering
\begin{tabular}{@{}lrrr@{}}
\toprule
\textbf{Model} & \textbf{Brier score} &  \textbf{Win Rate vs Best} & \textbf{Pairwise p-value}\\
\midrule 

OpenAI GPT-4o & 0.1485 & 0.4416 & <0.001 \\
Gemini 2.5 Pro & 0.1456 & 0.4099 & <0.001 \\
OpenAI GPT-5 & 0.1412 & 0.5115 & <0.001 \\
Qwen-32b & 0.1404 & 0.3155 & <0.001 \\
OpenAI o3 & 0.1242 & 0.4422 & 0.1561 \\
\textbf{Claude Sonnet 4} & 0.1195 & - & - \\
\bottomrule
\end{tabular}
\caption{Performance of various base models on \MarketLiquid}
\label{tab:market-results-other-models}    
\end{table}

As Table~\ref{tab:market-results-other-models} shows, Sonnet 4 and OpenAI o3 yield comparable performance on the \MarketLiquid; all other models perform significantly worse.\footnote{We find that Claude Opus 4 is prohibitively slow on our benchmarks and thus exclude it from our analysis.}  In addition to those presented above, we also find that OpenAI OSS-120b achieves a Brier score of 0.1076, with a win rate of 50\% against Sonnet 4. However, closer inspection reveals that this apparently state-of-the-art performance is a mirage --- the model is simply able to exploit the foreknowledge bias described in Section \Ref{sec:mitigation} more effectively than all other models we evaluate. Indeed, the analysis described in Section \Ref{sec:mitigation} reveals that 17.09\% of the news queries leak foreknowledge about the event in question. These problematic queries cumulatively affect 57\% of our benchmark, making it virtually impossible to evaluate the results. 

We default to OpenAI o3 as our base model unless otherwise specified. Unlike Sonnet 4, its pretraining cutoff (October 2023) is well before the events which appear in \forecastbench and \MarketLiquid, which precludes any form of pretraining lookahead bias on either benchmark \citep{sarkar2024lookahead}.

 \refstepcounter{section}  
\section*{C. Sources of Foreknowledge Bias}
\addcontentsline{toc}{subsection}{C. Sources of Foreknowledge Bias}
\label{sec:sources-foreknowledge}
\setcounter{subsection}{0}  % Reset subsection counter                     

Through extensive analysis of search API responses across \ProviderA and \ProviderB, we find that the primary mechanism for introducing foreknowledge bias is responses which return web pages or news articles that are updated after initial publication. They can be updated because of (1) live data feeds embedded on the page and/or (2) web pages that are continuously updated by the publisher.

\textbf{Live data feeds that bypass API date filters.}
When search APIs crawl and summarize web pages, they often capture real-time widgets that display current data regardless of our specified date constraints. Consider this concrete example: we called the \ProviderA API requesting information about Tapestry Inc. (TPR) stock with an explicit July 21, 2024 cutoff date. Despite this restriction, the API returned:

\begin{quote}
\small
``As of July 23, 2025, TPR's price was approximately \$108.72, meaning on July 21, 2024, the price was likely lower but near analyst forecasts of around \$91.82...''
\end{quote}

The API pulled this future price from a live widget on TipRanks, defeating the purpose of our temporal restriction. This occurs because the search APIs retrieve the current version of web pages, which often embed dynamically updating content.

\textbf{Search APIs return the current versions of continuously updated pages.} Wikipedia and similar platforms pose a distinct challenge: search APIs typically return the latest version of these pages, not a historical snapshot, even when a past date is specified.

\textbf{Search APIs returning current versions of continuously updated pages.} Wikipedia and similar platforms present a different challenge: when search APIs retrieve these pages, they get the most recent version rather than a historical snapshot, even when we specify a past date.

For instance, when we use the \ProviderA API to ask whether professional chess player Arjun Erigaisi will maintain his current ELO ranking, with an explicit July 21, 2024 cutoff date, the API returned:

\begin{quote}
\small
``...By July 2024, he had entered the top 5 Open with a rating of 2778 after gaining 17 points. As of December 2024, he became the 16th player in history to cross the 2800 rating mark... The Tata Steel Chess Tournament 2025, where he finished 10th, might affect his standings slightly...''
\end{quote}

Despite the intended date cutoff, the API revealed that Erigaisi crossed the 2800 threshold in December 2024, and even provided results from a January 2025 tournament. This occurred because the API pulled from a current, recently updated Wikipedia page, rather than from a historical snapshot.

 \refstepcounter{section}  
\section*{D. Foreknowledge Analysis Prompt}
\addcontentsline{toc}{subsection}{D. Foreknowledge Analysis Prompt}
\label{app:foreknowledge-prompt}
\setcounter{subsection}{0}  % Reset subsection counter                     

\begin{tcolorbox}[
  colback=lightgray,
  colframe=darkblue,
  arc=3pt,
  boxrule=1pt,
  fonttitle=\bfseries,
  title=Foreknowledge Analysis Prompt - Overview
]
\small
\textbf{You are analyzing a forecast question where a language model made a very confident prediction that turned out to be correct. Your task is to determine if there's evidence that the model may have used FOREKNOWLEDGE of the outcome rather than making a genuine prediction.}

\vspace{0.5em}
\textbf{QUESTION:} \texttt{\{question\}}

\textbf{MODEL'S KNOWLEDGE CUTOFF:} \texttt{\{news\_end\_date\}}

\textbf{RESOLUTION DATE:} \texttt{\{resolution\_date\}}

\textbf{CORRECT ANSWER:} \texttt{\{resolution\}}

\textbf{MODEL:} \texttt{\{model\_name\}}

\textbf{MODEL'S PROBABILITY ASSIGNMENT:} \texttt{\{probability:.4f\}} (very confident)

\textbf{MODEL'S REASONING AND DATA:} \texttt{\{model\_output\}}
\end{tcolorbox}

\begin{tcolorbox}[
  colback=lightgray,
  colframe=darkblue,
  arc=3pt,
  boxrule=1pt,
  fonttitle=\bfseries,
  title=Analysis Criteria
]
\small
\textbf{Please analyze the model's reasoning to determine if there's evidence of foreknowledge. Look for:}

\begin{enumerate}
\item \textbf{CLEAR EVIDENCE of outcome knowledge:} Does the reasoning explicitly reference the actual outcome, result, or resolution that occurred?
\item \textbf{Past-tense language about future events:} Does the model speak about events after the knowledge cutoff date as if they already happened? Recall that the model's knowledge cutoff is \texttt{\{news\_end\_date\}}, whereas the resolution date is \texttt{\{resolution\_date\}}.
\item \textbf{Explicit outcome statements:} Does the model state facts about what actually occurred rather than what might occur?
\item \textbf{Access to post-event information:} Does the reasoning reference news or information that could only be available after the event resolved?
\end{enumerate}

\vspace{0.5em}
\textbf{IMPORTANT:} We are NOT penalizing correct predictions or good forecasting. We are only looking for cases where the model clearly had access to information about the actual outcome.

\vspace{0.5em}
Recall that you have access to a search tool through which you can retrieve up to date information. Please feel free to use this search tool to investigate whether particular statements made by the model indicate knowledge of events after the knowledge cutoff date.
\end{tcolorbox}

\begin{tcolorbox}[
  colback=lightgray,
  colframe=darkblue,
  arc=3pt,
  boxrule=1pt,
  fonttitle=\bfseries,
  title=Required Response Format
]
\small
\textbf{Return your analysis in this JSON format:}
\begin{verbatim}
{
  "has_foreknowledge": boolean (true if clear evidence of 
                                foreknowledge exists),
  "confidence_level": "high" | "medium" | "low" (how confident you are 
                      in the foreknowledge assessment),
  "evidence_quotes": ["exact quote 1 showing foreknowledge", 
                      "exact quote 2", ...],
  "evidence_explanation": "detailed explanation of why the evidence 
                          suggests foreknowledge",
  "legitimate_reasoning": boolean (true if the reasoning could be 
                          legitimate forecasting without foreknowledge),
  "key_indicators": ["indicator 1", "indicator 2", ...] (specific signs 
                    of foreknowledge),
  "overall_assessment": "brief summary of whether this appears to be 
                        genuine forecasting or potential cheating"
}
\end{verbatim}

\textbf{Be very careful to distinguish between:}
\begin{itemize}
\item Strong forecasting based on available information (NOT foreknowledge)
\item Clear evidence of knowing the actual outcome (IS foreknowledge)
\end{itemize}
\end{tcolorbox}

 \refstepcounter{section}  
\section*{E. Further Search Details and Results}
\addcontentsline{toc}{subsection}{E.  Further Search Details and Results}
\label{app:blocklist}
\setcounter{subsection}{0}  % Reset subsection counter                     

The following URLs were excluded due to frequent post–cutoff leakage via dynamic widgets or continuously updated content (see Sec.~\ref{sec:mitigation}).

{\small
\begingroup
\def\UrlBreaks{\do\/\do-\do\_\do.\do?\do&}%
\begin{multicols}{2}
\begin{itemize}[leftmargin=*, itemsep=0.2ex, parsep=0ex, topsep=0.2ex]
\item \url{https://en.wikipedia.org/wiki/FIDE_rankings}
\item \url{https://en.wikipedia.org/wiki/Sarasadat_Khademalsharieh}
\item \url{https://en.wikipedia.org/wiki/Nodirbek_Abdusattorov}
\item \url{https://weatherspark.com/h/m/56493/2024/7/Historical-Weather-in-July-2024-in-Strasbourg-France}
\item \url{https://weatherspark.com/h/m/147731/2024/7/Historical-Weather-in-July-2024-at-Brest-Brittany-France}
\item \url{https://weatherspark.com/h/m/50604/2024/7/Historical-Weather-in-July-2024-in-Lyon-France}
\item \url{https://www.macrotrends.net/stocks/charts/HUM/humana/stock-price-history}
\item \url{https://www.nasdaq.com/market-activity/stocks/tsco}
\item \url{https://en.wikipedia.org/wiki/Viswanathan_Anand}
\item \url{https://www.historique-meteo.net/caraibes/guadeloupe/pointe-a-pitre/2024/07/}
\end{itemize}
\end{multicols}
\endgroup
}

In Table \Ref{tab:prevalence_estimate}, We include additional results for the LLM-as-a-judge described in Section \Ref{sec:mitigation}. These estimates suggest that $\sim 1.65\%$ of search results include foreknowledge bias. 

\begin{table}[htbp]
\centering
\begin{tabular}{@{}lcc@{}}
\toprule
\textbf{Quantity (N=4411)} & \textbf{Estimate} & \textbf{Notes} \\
\midrule
Flags (judge positive)         & 236  & 5.35\% of responses \\
Estimated True Positives (TP)  & $\approx 47$   & $= \mathrm{Precision} \times \text{flags}$ \\
Estimated False Positives (FP) & $\approx 189$  & $=$ flags $-$ TP \\
Estimated False Negatives (FN) & $\approx 26$   & from $R{=}TP/(TP{+}FN)$ \\
Estimated True Negatives (TN)  & $\approx 4149$ & $=$ clean $-$ FN \\
Estimated Foreknowledge Bias       & $\approx 1.65\%$   & $(TP{+}FN)/N$ \\
\bottomrule
\end{tabular}
\caption{Estimating the true rate of foreknowledge bias from the 5.35\% of flagged search results. Precision and recall estimates are derived from extensive manual validation of flagged responses, as described in Section \Ref{sec:mitigation}}
\label{tab:prevalence_estimate}
\end{table}

 \refstepcounter{section}  
\section*{F. \MarketLiquid Dataset Construction}
\addcontentsline{toc}{subsection}{F.  \MarketLiquid Dataset Construction}
\label{app:market}
\setcounter{subsection}{0}  % Reset subsection counter                     

In this section, we describe our prompts for rewording prediction market questions and our method of selecting date cutoffs. The date cutoffs depend on the length of time the market was open and were selected in consultation with a domain expert.

We first show the prompt used to reword questions:

\begin{tcolorbox}[
  colback=lightgray,
  colframe=darkblue,
  arc=3pt,
  boxrule=1pt,
  fonttitle=\bfseries,
  title=Question rewording prompt
]
The following information is from a prediction market where the title is templatized and may be missing information.\\
The rules contain more information that should be enough to fully resolve the market.\\
\\
Title: \{title\} \\
Rules: \{rules\} \\
\\
Please output a single yes or no question that encompasses the title and rules.\\
When possible, keep the phrasing as close to the title as possible, but ensure the question is a natural sentence.\\
ONLY output the question, do not output anything else.
\end{tcolorbox}

\newpage

Next, we show the pseudocode used to select the date cutoffs:

\begin{lstlisting}[
  language=Python,
  basicstyle=\ttfamily\small,
  keywordstyle=\color{blue},
  stringstyle=\color{orange},
  commentstyle=\color{gray},
  showstringspaces=false,
  frame=single,
  breaklines=true
]
open_time = ...
close_time = ...
diff = close_time - open_time
if diff.days > 91:
    days_from_close = [1, 7, 30, 60, 90]
elif diff.days > 30:
    days_from_close = [1, 7, 14, 21, diff.days]
elif diff.days > 7:
    days_from_close = [1, 3, 5, 7, diff.days]
else:
    # Invalid
\end{lstlisting}

 \refstepcounter{section}  
\section*{G. LLMs Attenuate Forecasts}
\addcontentsline{toc}{subsection}{G.  LLMs Attenuate Forecasts}
\setcounter{subsection}{0}  % Reset subsection counter                     

\subsection*{G.1 Examples of Hedging Behavior}
\addcontentsline{toc}{subsubsection}{G.1 Examples of Hedging Behavior}
\label{app:hedging}
\begin{tcolorbox}[
  colback=lightgray,
  colframe=darkblue,
  arc=3pt,
  boxrule=1pt,
  fonttitle=\bfseries,
  title=Partial Reasoning Trace Showing Hedging,
]
\small

\vspace{0.5em}
\textbf{QUESTION:} \texttt{Will the total return of the ICE BofA US Corporate Index, which tracks the performance of corporate debt issued in the US domestic market, have increased by 2024-08-20 as compared to its value on 2024-07-21?}

\textbf{MODEL'S KNOWLEDGE CUTOFF:} \texttt{2024-07-21}

\textbf{RESOLUTION DATE:} \texttt{2024-08-20}

\textbf{CORRECT ANSWER:} \texttt{1.0}

\textbf{MODEL:} \texttt{o3}

\textbf{MODEL'S PROBABILITY ASSIGNMENT:} \texttt{0.6} 

\textbf{MODEL'S REASONING AND DATA:} \texttt{The historical base rate of one-month positive returns for the ICE BofA US Corporate Index (or close analogues) is roughly 61\%.  Current carry does give a small edge, but markets already discount the possibility of a September rate cut; the asymmetric risk is that the Fed sounds ‘higher-for-longer,’ which could push yields up more than the carry buffer.  Summer illiquidity and election/geopolitical headlines further skew the distribution to the downside.  These considerations trim the edge implied by the initial 0.62 estimate back to an even 0.60, keeping the forecast slightly above the base rate but acknowledging that the upside case is already partly priced in.}
\end{tcolorbox}

\begin{tcolorbox}[
  colback=lightgray,
  colframe=darkblue,
  arc=3pt,
  boxrule=1pt,
  fonttitle=\bfseries,
  title=Partial Reasoning Trace Showing Hedging,
]
\small

\vspace{0.5em}
\textbf{QUESTION:} \texttt{According to Wikipedia, will César Cielo still hold the world record for 50m freestyle in long course (50 metres) swim?}

\textbf{MODEL'S KNOWLEDGE CUTOFF:} \texttt{2024-07-21}

\textbf{RESOLUTION DATE:} \texttt{2024-08-20}

\textbf{CORRECT ANSWER:} \texttt{1.0}

\textbf{MODEL:} \texttt{o3}

\textbf{MODEL'S PROBABILITY ASSIGNMENT:} \texttt{0.97} 

\textbf{MODEL'S REASONING AND DATA:} \texttt{The 14-year survival of Cielo’s mark implies a weekly break rate of roughly 0.13\%, suggesting >99\% chance the record survives any random week.  However, records are disproportionately broken at major meets and during taper periods; the Paris pool opens for training this week, creating a thin tail risk of a sanctioned time-trial or relay lead-off. We also add a small ‘Wikipedia-update’ risk. To hedge these low-but-non-zero unknowns, we shave the probability from 98\% to 97\%.}
\end{tcolorbox}

\subsection*{G.2 Platt Scaling is Generalized Log Odds Extremization}
\addcontentsline{toc}{subsubsection}{G.2  Platt Scaling is Generalized Log Odds Extremization}
\label{app:platt-extreme}

Log odds extremization as presented in \cite{tetlock-extreme-paper} is a special application of Platt scaling. In particular, log odds aggregation is the Platt scaling of the geometric mean of forecasts.

First we show that Equation 1 from \cite{tetlock-extreme-paper} is equivalent to Platt scaling. This allows us to express Platt scaling in a simpler form. That is, we show that $\frac{1}{1 + \exp(-a\log \frac{p}{1-p})}$ and $\frac{p^a}{p^a + (1-p)^a}$ are equivalent expressions. Starting with $\frac{p^a}{p^a + (1-p)^a}$ and $p = f(x)$ for some real valued function:

\begin{align}
    \frac{p^a}{p^a + (1-p)^a} &= \frac{1}{1+\left(\frac{1-p}{p}\right)^a} \\
    &= \frac{1}{1+e^{\alpha\log\left(\frac{1-p}{p}\right)}} \\
    &= \frac{1}{1+e^{-\alpha\log\left(\frac{p}{1-p}\right)}} \\
    &= \text{sigmoid}\left(\alpha\log\left(\frac{p}{1-p}\right)\right)
\end{align}

A more general version of Platt scaling exists, where the $d$ parameter controls the sigmoidal curve to move left and right along the domain. Here, we show that the $d$ parameter manifests as a $\log d$ term in the sigmoid of Platt scaling.

\begin{align}
    \frac{dp^a}{dp^a + (1-p)^a} &= \frac{1}{1+\frac{1}{d}\left(\frac{1-p}{p}\right)^a} \\
    &= \frac{1}{1+e^{\alpha\log\left(\frac{1-p}{p}\right) - \log d}} \\
    &= \frac{1}{1+e^{-\alpha\log\left(\frac{p}{1-p}\right)-\log d}} \\
    &= \text{sigmoid}\left(\alpha\log\left(\frac{p}{1-p}\right) + \log d\right)
\end{align}

With this connection to Platt scaling in mind, we show that Equation \ref{eqn:extremization} is equivalent to Platt scaling of the geometric mean with a scaling parameter $a$ and no shift parameter i.e. $d=1$.

Beginning with log odds as prescribed in Equation \ref{eqn:extremization}, $\log \frac{\hat{p}}{1-\hat{p}}$ where $\hat{p}$ is the aggregated scaled forecast, we can impute $\hat{p}$ by applying a $\text{sigmoid}(\cdot)$. Then simplifying gives: 

\begin{align*}
    \hat{p} &= \text{sigmoid}\left(\log \frac{\hat{p}}{1-\hat{p}}\right) \\
    &=\text{sigmoid}\left(a\frac{1}{n}\sum_{i=1}^n \log \frac{p_i}{1-p_i}\right) \\ 
    &= \text{sigmoid}\left(\frac{1}{n}\sum_{i=1}^n \left(\log \frac{p_i}{1-p_i}\right)^a\right) \\ 
    &= \frac{1}{1+\exp\left(-\sum_{i=1}^n \left(\log \frac{p_i}{1-p_i}\right)^{a/n}\right)} \\
    &= \frac{1}{1+\frac{\prod_{i=1}^n (1-p_i)^{a/n}}{\prod_{i=1}^n p_i^{a/n}}} \\
    &= \frac{1}{1+\frac{\prod_{i=1}^n (1-p_i)^{a/n}}{\prod_{i=1}^n p_i^{a/n}}} \\
    &= \frac{\prod_{i=1}^n p_i^{a/n}}{\prod_{i=1}^n p_i^{a/n}+\prod_{i=1}^n (1-p_i)^{a/n}} \\
    &= \frac{\left(\prod_{i=1}^n p_i^{1/n}\right)^a}{\left(\prod_{i=1}^n p_i^{1/n}\right)^a+\left(\prod_{i=1}^n (1-p_i)^{1/n}\right)^a}
\end{align*}

Thus the log odds extremized score $\hat{p}$ is a special application of Platt scaling. Specifically, it is the Platt scaling of the geometric mean of each $p_i$ and $1-p_i$, with Platt scaling parameter $a$ and $d=1$.

\end{document}